\documentclass[english,11pt]{elsarticle} 
\usepackage[pdftex]{color}
\usepackage[font=footnotesize,labelfont=bf]{caption}
\usepackage[font=footnotesize,labelfont=bf]{subcaption}

\usepackage[T1]{fontenc}
\usepackage[latin9]{inputenc} 
\usepackage{amsmath}
\usepackage{amsthm}
\usepackage{amssymb}
\usepackage{amsfonts}
\usepackage{bbm}
\usepackage{graphicx}
\usepackage{epstopdf}
\usepackage[pdftex,colorlinks,linkcolor={blue},citecolor={blue}]{hyperref}

\usepackage{pgfplots}
\pgfplotsset{compat=newest}
\usetikzlibrary{plotmarks}
\usetikzlibrary{arrows.meta}
\usepgfplotslibrary{patchplots}
\usepackage{grffile}

\usepackage{epsfig}
\usepackage[font=footnotesize,labelfont=bf]{caption}
\usepackage[font=footnotesize,labelfont=bf]{subcaption}

\usepackage{amssymb,amsmath,amsfonts,layout,graphicx}
\usepackage{makeidx}
\usepackage{babel}
\usepackage{tikz}
\usepackage{sublabel}
\usepackage{cases}
\usepackage{algorithm}
\usepackage{algorithmic}
\usepackage[font=small,labelfont=bf]{caption}

\usepackage
[
        letterpaper,% other options: a3paper, a5paper, etc
        left=1.91cm,
        right=1.91cm,
        top=1.91cm,
        bottom=2cm,
]
{geometry}

\newtheorem{assumption}{Assumption}

\newtheorem{remark}{Remark}

\newcommand{\x}{{\bf x}}

\begin{document}
\begin{frontmatter}

\title{A Queuing Approach to Parking:\\ Modeling, Verification, and Prediction}
%\tnotetext[mytitlenote]{Fully documented templates are available in the elsarticle package on \href{http://www.ctan.org/tex-archive/macros/latex/contrib/elsarticle}{CTAN}.}

% Group authors per affiliation:
\author[1staddress]{Hamidreza Tavafoghi}
\ead{tavaf@berkeley.edu}
\author[1staddress,2ndaddress]{Kameshwar Poolla}
\ead{poolla@berkeley.edu}
\author[2ndaddress]{Pravin Varaiya}
\ead{varaiya@berkeley.edu}

\address[1staddress]{Department of Mechanical Engineering, University of California, Berkeley }
\address[2ndaddress]{Department of Electrical Engineering and Computer Science, University of California, Berkeley}
%\fntext[myfootnote]{Since 1880.}

%% or include affiliations in footnotes:
%\author[mymainaddress,mysecondaryaddress]{Elsevier Inc}
%\ead[url]{www.elsevier.com}

%\author[mysecondaryaddress]{Global Customer Service\corref{mycorrespondingauthor}}
%\cortext[mycorrespondingauthor]{Corresponding author}
%\ead{support@elsevier.com}

%\address[mymainaddress]{1600 John F Kennedy Boulevard, Philadelphia}
%\address[mysecondaryaddress]{360 Park Avenue South, New York}

\begin{abstract}
We present a queuing model of parking dynamics and  a model-based prediction method to provide real-time probabilistic forecasts of future parking occupancy.  The  queuing model has non-homogeneous arrival rate and time-varying service time distribution. All statistical assumptions of the model are verified using data from 29 truck parking locations, each with between 55 and 299  parking spots. For each location and each spot the data specifies the arrival and departure times of a truck, for 16 months of operation.
The modeling framework presented in this paper provides empirical support for   queuing models adopted in  many theoretical studies and policy designs. We discuss how our framework can be used to study  parking problems in different environments. Based on the queuing model, we propose two prediction methods, a microscopic method and a macroscopic method, that  provide a real-time probabilistic forecast of parking occupancy for an arbitrary forecast horizon. These model-based methods convert a probabilistic forecast problem into a parameter estimation problem that can be tackled using classical estimation methods such as regressions or pure machine learning algorithms.  We characterize a lower bound for an arbitrary real-time prediction algorithm. 
We evaluate the performance of these methods using the truck data comparing the outcomes of their implementations with other model-based and model-free methods proposed in the literature.
\end{abstract}

%\begin{keyword}
%\end{keyword}

\end{frontmatter}

\section{Introduction}

Every vehicle trip involves parking. It has been suggested that in congested downtowns, it takes between $3.5$ to $14$ minutes to find a parking spot and that between $8$ to $74$ percent of vehcles are cruising in search of  parking  \cite{shoup2006cruising,inci2015review,guiffre2006novel}.
Parking areas take up to an equivalent of $30\%$ land  in many big cities like San Francisco or Toronto\footnote{Most of the parking spaces in central business districts (CBD) are underground or vertical structures. The parking coverage is calculated based on how much land parking spaces would occupy if they were spread horizontally over a surface lot.}; this number reaches $81\%$ in automobile-dependent cities like Los Angeles \cite{manville2005parking}. Therefore, an understanding of parking dynamics is crucial for a cost-efficient sizing and operation of parking facilities.

In recent years, thanks to the increasing deployment of stationary sensors \cite{sfpark}, mobile sensing \cite{mathur2010parknet}, and camera-based detection \cite{bulan2013video}, it is possible to create a real-time map of parking occupancy with high spatio-temporal resolution; see \cite{lin2017survey} for a survey. Moreover, with online information systems that share this real-time parking information with drivers, it becomes possible to inform drivers and pro-actively influence their parking choices through  such means as dynamic pricing and online reservation systems. To fully utilize the potential of these emerging technologies we need a parking model to analyze the impact of different policies and form real-time parking prediction to inform drivers.

There is a growing literature on modeling and analysis of parking. Queuing models have been the primary  choice in many theoretical and numerical parking studies.  Arnott \textit{et al.} \cite{arnott1999modeling} present an economic model of parking based on an approximation of a queuing model. The work in \cite{portilla2009using} uses a $M/M/\infty$ queuing model to study the effect of cruising on congestion. Using a $M/M/C$ model, the authors in \cite{millard2014curb} study the impact of variable pricing on  parking occupancy and cruising in San Francisco. Assuming stationary arrivals and departures, the authors in \cite{arnott2006integrated,arnott2009downtown,arnott2013bathtub} propose a \textit{bathtub model} to capture the effect of cruising on traffic congestion and study  parking pricing. Similarly, \cite{larson2007congestion} considers a parking pricing problem using an approximate queuing model assuming a sufficiently large population. More recently, the authors in \cite{dowling2019modeling} study  curbside parking  using a model of networked queues with limited capacities.
While queuing models are widely used to study parking, it is difficult to find any empirical study that  verifies the   statistical assumptions about arrival and departure processes that underlie these queuing models.

The first contribution of this paper is to address this verification question. We present a queuing model for parking with formal statistical validation of all of its assumptions using real data. We consider a queuing model where the arrivals follow a non-homogeneous Poisson process (NHPP), and parking times are distributed according to a time-varying probability distribution. The methodology we use to verify the assumptions of the queuing model suggests a general approach that can be applied using real datasets from other parking environments.

The second contribution consists of two  methods that provide a probabilistic forecast of parking occupancy based on the characterized queuing model. The first method follows a microscopic approach by considering each parking spot separately, while the second  uses a macroscopic approach treating a collection of parking spaces on an aggregate level. We use learning algorithms to learn/predict parameters of the prediction model. Using the queuing framework, we characterize an upper bound on the accuracy of an arbitrary parking prediction method. We demonstrate the performance of the prediction methods we propose and compare them against the described bound. 

We use data from truck parking lots operated by Pilot Flying J. company in  locations across California, Oregon, and Washington. The data helps us to develop and verify a \textit{basic model} avoiding a few additional complexities that are present in a congested CBD including potential interdependencies between parking demands in neighboring regions and censored/unobserved demand whenever a parking lot is full. We discuss how the basic model   can be extended to study parking problems in CBD addressing these additional complexities.

\section{Related Literature}\label{sec:relatedlit}

There is a growing literature on prediction of parking occupancy. In contrast to our results, most   prediction methods only provide a point-wise prediction of parking occupancy. This literature can be classified into two main categories : (i) a class of model-based studies that consider a queuing model or a variation of it for analysis or prediction  \cite{caliskan2007predicting,klappenecker2014finding,pullola2007towards,chen2013uncertainty,monteiro2018street,rajabioun2013intelligent,xiao2018how}, and (ii) a class of studies that use a generic machine learning algorithm for prediction \cite{zheng2015parking,chen2014parking,richter2014temporal,fiez2018data,vlahogianni2016real,demisch2016demand,rajabioun2015street,ji2014short,park1999spectral,camero2018evolutionary,vanajakshi2005estimation,okutani1984dynamic,ji2010applied,rice2004simple,kwon2000day,williams2003modeling,van1996combining,vlahogianni2004short,smith2002comparison,bock2017data}.

Many prediction methods have been developed based on a queuing model. The authors in \cite{caliskan2007predicting,klappenecker2014finding} propose a prediction method by considering an $M/M/C$ queuing model and perform a numerical study. The works in \cite{pullola2007towards,chen2013uncertainty,monteiro2018street,rajabioun2013intelligent} propose prediction methods assuming  arrivals to parking, departures from parking, or available parking spots follow a NHPP. While some of these works do not explicitly use a queuing model to describe their prediction algorithms, their results can be viewed as based on various approximations of a queuing model for large populations.      	
Specifically, the authors in \cite{pullola2007towards} propose a prediction method by assuming the parking availability follows a NHPP and demonstrate the performance of their approach through simulations. In  \cite{chen2013uncertainty}, the authors model cars leaving a parking lot as a NHPP and use it to predict the waiting time when a parking lot is full. In \cite{monteiro2018street,rajabioun2013intelligent}, the authors assume that both arrivals and departure to/from a parking lot are NHPP, and propose a short-term prediction method accordingly. The prediction method developed in \cite{xiao2018how} is the closest to our appraoch in this paper. The authors in \cite{xiao2018how} considers an $M/M/C$ queuing model to present a prediction framework for parking occupancy and demonstrate their results using real data. However, the authors in \cite{xiao2018how} do not provide a formal statistical validation of their model assumptions. Moreover, by assuming a $M/M/C$ queuing model the authors restrict their model to environments where arrival rates are fixed over time and service times are exponentially distributed. Neither   assumption is likely to be true for real data. 
Specifically, the exponential distribution assumption made in \cite{pullola2007towards,chen2013uncertainty,monteiro2018street,rajabioun2013intelligent,xiao2018how} implies that the probability distribution of the additional time a vehicle will keep occupying a parking spot is independent of time it has already occupied the spot. This assumption is hard to justify in many situations where parking demand depends on time, e.g. during working hours in CBD, or where parking spots have a  time limit.
As we show when service times are non-exponentially distributed utilizing the current age of each parked vehicle can improve the accuracy for short-term prediction; this is specially true in urban areas where there are parking time limits or curbside parking with metering.

An alternative to the model-based approach  is to use data-driven machine learning algorithms to  predict  future occupancy directly from the past observed data with no assumption about the underlying model. The authors in \cite{rajabioun2015street} propose a multivariate autoregressive model taking into account spatial and temporal correlations to predict parking
availability in different parking locations in the short term (30 minutes). The work in \cite{richter2014temporal} uses spatio-temporal clustering to predict parking availability using  data from San Fransisco. The authors in \cite{fiez2018data} use a Gaussian mixture model to predict parking demand in Seattle. The authors in \cite{chen2013uncertainty} investigate four parking prediction methods, namely, autoregressive integrated moving average (ARIMA), linear regression, support vector regression (SVR), and feed-forward neural network. Similarly, the authors in  \cite{zheng2015parking} analyze three prediction methods including regression tree, SVR, and feed-forward neural network. Different types of neural networks have been used for prediction including deep neural networks \cite{camero2018evolutionary}, wavelet neural networks \cite{ji2014short}, spectral basis neural networks \cite{park1999spectral}, and  feed-forward neural network with a sliding window \cite{vlahogianni2016real}.
While the use of model-free algorithms for prediction may often provide a better performance in comparison to model-based approaches thanks to a richer prediction function space they provide, they suffer from two main drawbacks. First, a model-free algorithm generally has many hyper-parameters that need to be individually estimated/tuned for every dataset, e.g. the number of layers, size of each layer, and learning rate in a neural network. As a result, a model developed based on a specific dataset has little applicability for a new dataset coming from a different environment. 
Second, these methods have limited value in exploratory studies and policy analysis. This is because models resulting from these approaches are not interpretable in terms of a language descriptive of a broader transportation system and provide limited connection to the existing literature on other aspects of transportation networks

In this paper, we present a queuing model whose underlying statistical assumptions  are verified empirically. We develop two model-based methods for a probabilistic prediction of parking occupancy. We also use machine learning algorithms to learn/predict parameters of the queuing model that we use for parking prediction. Therefore, we utilize the potential strength of machine learning algorithms while we also benefit from the implicit structure in a model-based approach.

\section{Outline}

The rest of the paper is organized as follows. In Section \ref{sec:data}, we  describe the real dataset  used in this paper.
In Section \ref{sec:model-description}, we present the basic queuing model we use to explain the observed data and discuss the assumptions underlying the model. We provide a statistical verification of these assumptions in Section \ref{sec:model-verification}.
In Section \ref{sec:analysis}, we provide a statistical analysis of the  two main components of the model, namely the arrivals and service times. 
In Section \ref{sec:analysis-arima}, to capture the time variation of the service time distribution function, we present a parametrized partition of arrivals into a set of heterogeneous populations and expand the basic model accordingly. 
In Section \ref{sec:prediction}, we present our methods for the prediction of parking occupancy based on the model and analysis described in Sections \ref{sec:model-description}-\ref{sec:analysis-arima}. We evaluate the performance of the proposed methods and discuss their advantage relative to other model-free and model-based prediction methods. 
We conclude in Section \ref{sec:conclusion} and briefly discuss a few possible directions for future work.

\begin{figure}[t!]
	\begin{minipage}{\textwidth}
		\centering
		\includegraphics[width=0.95\textwidth]{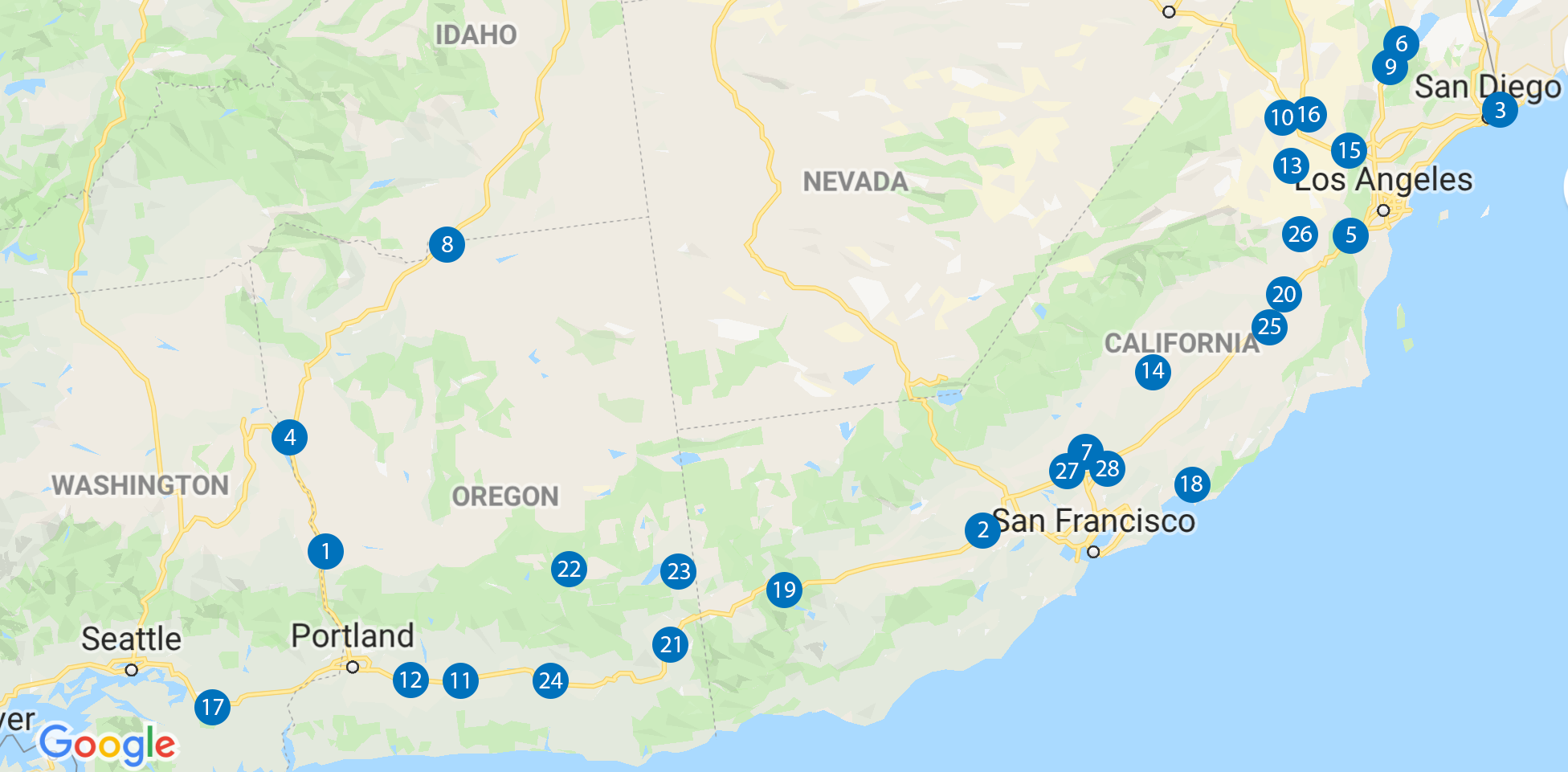}
		\caption{The location of parking lots across the west coast of the united states}
		\label{fig:map}
	\end{minipage}%x
\end{figure}%

\section{Data Description}\label{sec:data}

The real data we use in this study are obtained from $29$ truck parking lots operated by Pilot Flying J. across the west coast of the United States; see Figure \ref{fig:map}. Each parking lot is located in a rest area that houses a convenience store, gas station, restaurants, and possibly a few other stores such as repair shops.
Every parking lot  has between $55$ to $299$ parking spots with an average of $129$ spots per location; we only consider parking spots that are occupied at least twice during a year. Each parking spot is equipped with a sensor that detects an arrival or a departure of a vehicle, and communicates the time of each event with a database in real-time. The data used in this study are from observations made between August 01, 2016 to June 06, 2018; the data for some parking lots cover a smaller time interval as their sensing systems became operational later than August 01, 2016. There are a few ($\sim 0.5\%$) faulty measurements in the dataset for each parking lot, e.g. two arrivals with no departure in between. We keep this data  and add the necessary arrivals/departures entries setting their timestamps as the average time between the immediate events that occur before and after them. Figure \ref{fig:3day-variation} shows an example of parking occupancy variation during the span of three days for one of these parking lots.

\section{Basic Model}\label{sec:model-description}

\begin{figure}[t!]
	\begin{minipage}{0.42\textwidth}
		\centering
		\hspace*{-10pt}\includegraphics[width=0.90\textwidth,height=4.5cm]{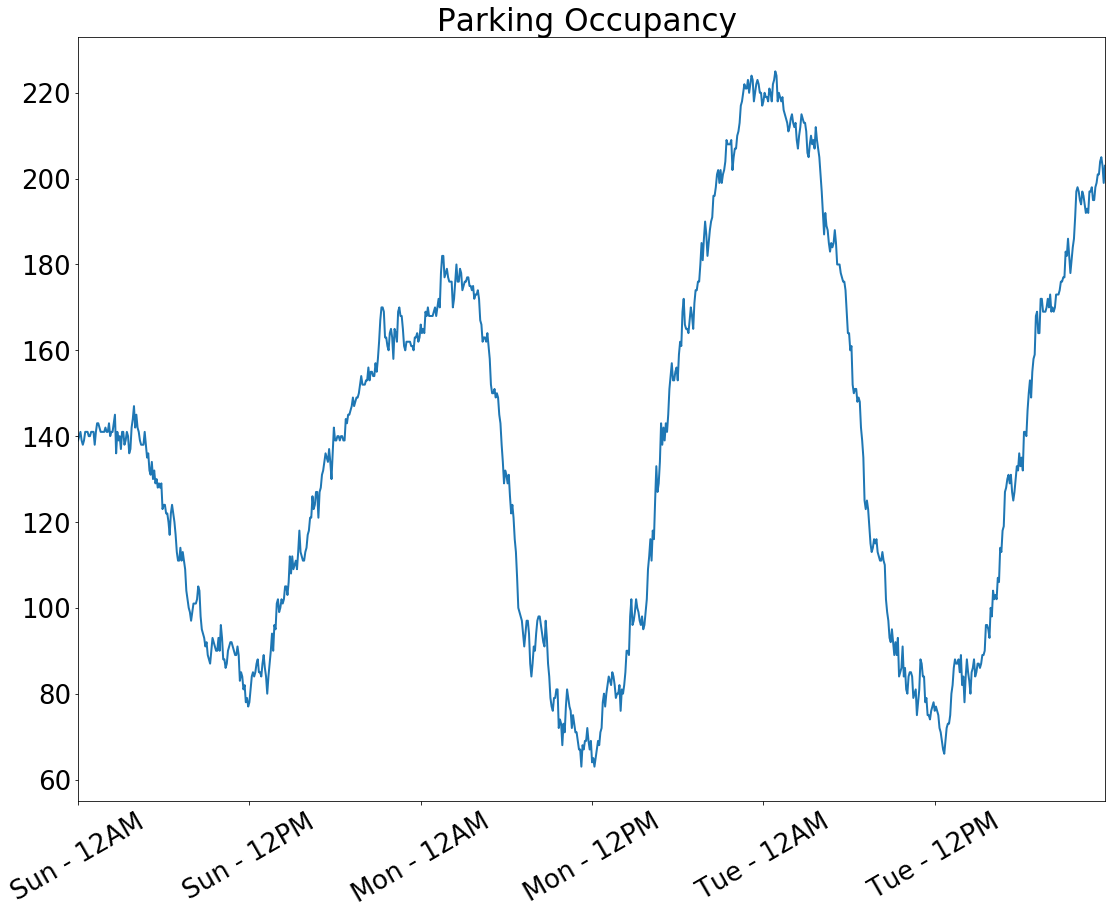}\vspace{5pt}
		\caption{Parking occupancy at location 20 between March 4 - 7, 2018.}
		\label{fig:3day-variation}
	\end{minipage}%
	\hspace*{10pt}
	\begin{minipage}{0.54\textwidth}
	\centering
	\includegraphics[width=\textwidth]{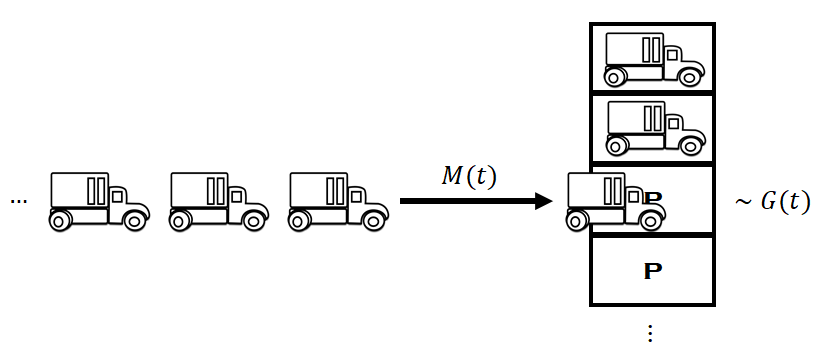}	
	\caption{$M(t)/G(t)/\infty$ queuing model for each parking lot}
	\label{fig:model}
	\end{minipage}
\end{figure}

Each parking lot is modeled separately.\footnote{We discuss in Section  \ref{sec:analysis-arima} the potential value in considering a networked model that captures all parking lots in one framework in order to exploit possible dependencies and correlation among different lots.} We consider an $M(t)/G(t)/\infty$ queuing model \cite{whitt2018time}  for each parking lot (Figure \ref{fig:model}). Vehicles arrive according to a non-homogeneous Poisson process (NHPP) at rate $M(t)$ vehicles per hour. Once a vehicle arrives at time $t$, it parks for a random time $S$  distributed according to a cumulative probability distribution (CDF) $G(t)$; we assume that, conditioned on the arrival time $t$, the amount of time each vehicle parks is independent of all  other vehicles' arrival times and parking durations. Following the terminology in the queuing literature, we refer to the time each truck parks as its \textit{service time}. We assume that there are an infinite number of parking spots at each parking lot, so no arriving truck finds the parking lot full. Below, we describe formally the assumptions we make for this queuing model.

%\subsection{Assumptions}

Starting from time $t=0$, we index vehicles entering the parking lot based on their arrival time (increasing in time) as vehicle $1,2,\cdots$ .  Let $\tau_i$ denote the arrival time of  vehicle $i$, and let $A(t):=\sum_{i=1}^\infty \mathbbm{1}_{\{\tau_i\leq t\}}$ be the number of vehicles that arrive up to  $t$. The first assumption we make in the $M(t)/G(t)/\infty$ queuing model is on the arrival process $A(t)$.

\begin{assumption}\label{assump-poisson}
	The arrival process $A(t)$ is a non-homogeneous Poisson process.
\end{assumption}

Assumption \ref{assump-poisson} is equivalent to requiring that for every  $t\geq0$, $A(t)$ is a Poisson random variable with mean $\mu_A(t)$ given by
\begin{align}
\mu_A(t)=\int_0^t M(s) ds,
\end{align}
where $M(t)$ denotes the time-dependent arrival rate at time $t$.

Let $S_i$ denote the (random) service time/parking duration for vehicle $i$. The second assumption concerns this service time.

\begin{assumption}\label{assump-independent}
	Service times $S_i$, $i=1,2,\cdots$ are independent random variables.
\end{assumption}

Assumption \ref{assump-independent} does not imply that service times $S_i$, $i=1,2,\cdots$  are identically distributed. Indeed, we consider the case where service time $S_i$ is distributed according to a time-varying CDF $G(\tau_i)$. Therefore to describe the $M(t)/G(t)/\infty$ queue we need to specify $M(t)$ and $G(t)$ for all $t\geq0$.

Here is the third assumption.
\begin{assumption}\label{assump-infinite}
	No arriving vehicle finds the parking lot full i.e. there are infinitely many parking spots.
\end{assumption}

Of course no real-world parking lot has an infinite number of spots. But if the occupancy level of a parking lot never reaches $100\%$, this is equivalent to having infinitely many spots. Given Assumption \ref{assump-infinite}, the number of occupied spots $N(t)$ at time $t$ is 
\begin{align}
N(t)=\sum_{i=1}^\infty \mathbbm{1}_{\{\tau_i\leq t\}}\mathbbm{1}_{\{\tau_i+S_i>t\}}, \label{eq:occupied-sum}
\end{align} 
which is the number of vehicles that arrive before time $t$ and leave  after time $t$. 

From Assumptions \ref{assump-poisson}-\ref{assump-infinite},  the number of occupied spots $N(t)$ is a Poisson random variable with mean
\begin{align}
	\mu_N(t)=\int_{0}^{\infty} M(t-s)(1-G(s;t))ds. \label{eq:occupied}
\end{align}
where $G(s;t)$ denotes the value of CDF $G(t)$ for service time $s$; see \cite{whitt2018time} for the proof. 

Thus the $M(t)/G(t)/\infty$ queuing model can provide a probabilistic forecast of parking occupancy over time if we characterize $M(t)$ and $G(t)$, or equivalently $\mu_N(t)$. Before proceeding with the analysis, we empirically verify the three assumptions underlying the model. We then analyze the  data in Sections \ref{sec:analysis} and \ref{sec:analysis-arima} based on the queuing model.

\section{Empirical Verification}\label{sec:model-verification}
Various queuing models have been used to study parking. However, to our knowledge, this work provides the first formal verification of the underlying statistical assumptions using real-world data. We proceed to verify Assumptions 1-3.

\subsection{No Censored Demand}
\begin{figure}[t!]
	\begin{minipage}{0.48\textwidth}
		\centering
		\hspace*{-20pt}\includegraphics[width=1.15\textwidth]{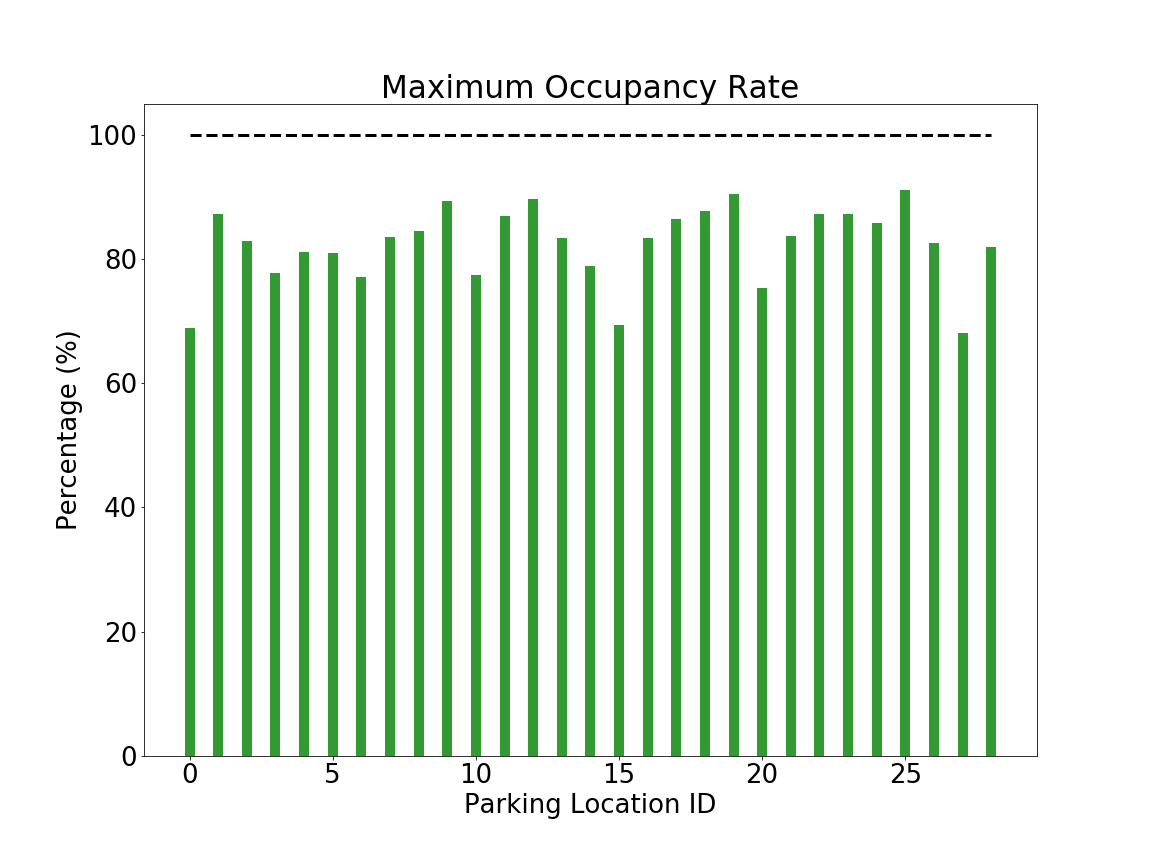}
		\caption{Maximum occupancy rate for each parking lot between November 27, 2017 and June 10, 2018.}
		\label{fig:assumpt-infinite}
	\end{minipage}%
	\hspace*{10pt}
	\begin{minipage}{0.48\textwidth}
		\centering
		\vspace*{10pt}
		\hspace*{-20pt}\includegraphics[width=1.15\textwidth]{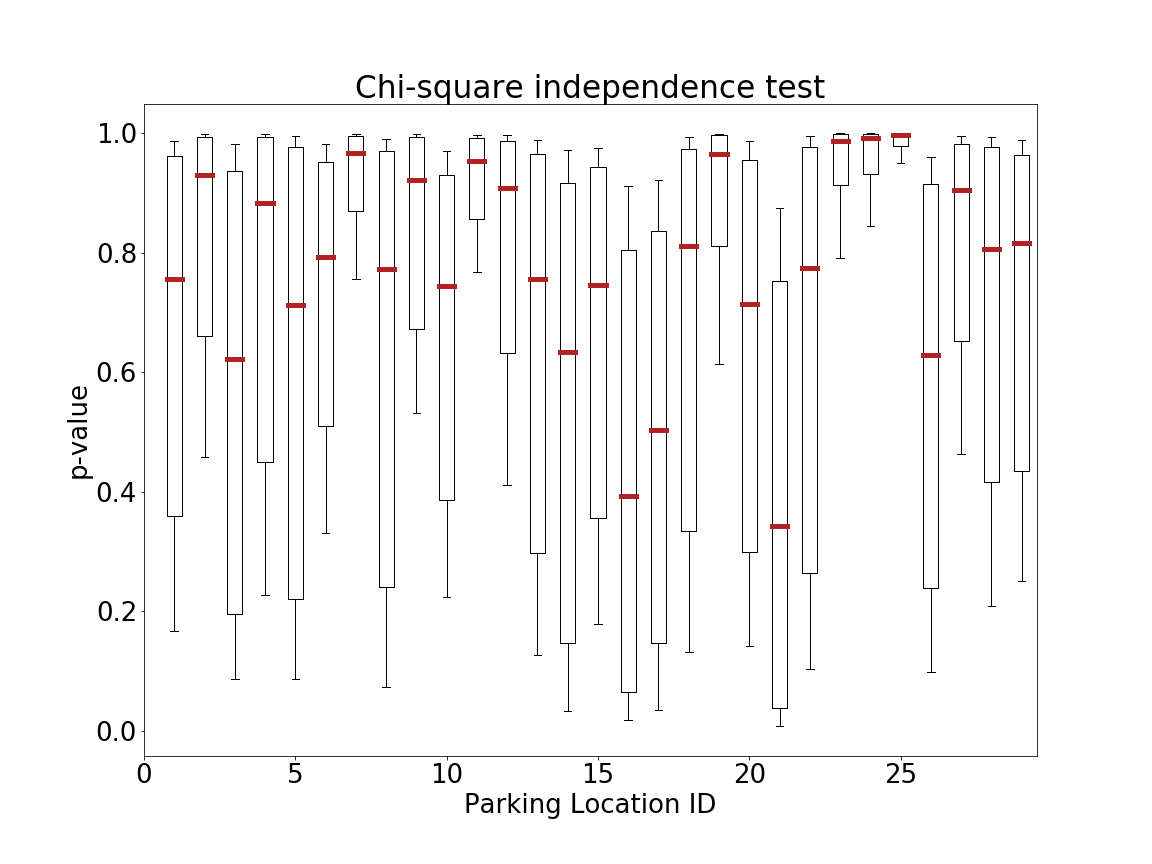}
		\caption{Chi-square independence test: Box plot of p-value for one-hour windows between March 1, 2018 and April 1, 2018, for every parking location.}
		\label{fig:assump-independence}
	\end{minipage}
\end{figure}

Assumption \ref{assump-infinite}  implies that all potential arrivals are \textit{observed},  so  there is no \textit{censored demand} because all parking spots are occupied at some time. In determining the occupancy of the parking lot, we only consider parking spots that have been occupied by a parked vehicle at least twice over the period of our study; otherwise, we consider them not suitable for parking and exclude them. 
Figure \ref{fig:assumpt-infinite} shows the maximum observed occupancy level at each parking lot between November 27, 2017 and June 10, 2018. Thus every parking lot always has  an empty suitable spot for an arriving vehicle, and so Assumption \ref{assump-infinite} holds.

\begin{remark}
In Sections \ref{sec:prediction} and \ref{sec:conclusion} we briefly discuss the case when a parking lot becomes full. We note that, for verifying the assumptions, one only need to consider data during  time periods when the parking lot is not full. However, for  prediction, we need to determine the actual arrival rate $M(t)$ even during  time intervals when the parking lot is full. This is a very challenging task and requires additional assumptions about driver behavior and parking demands during those times.\footnote{For instance, \cite{millard2014curb} assumes that  cruising drivers search until they find a parking spot.}
\end{remark}

\subsection{Independent Service Times}\label{sec:-verification-independence}

 In the $M(t)/G(t)/\infty$ model, service times are drawn from a time-varying probability distribution $G(t)$. Therefore, we cannot directly apply a statistical test for independence on all observed service times together since consecutive service times $S_i$ and $S_{i+1}$, $i\in\mathbb{N}$, are (indirectly) correlated through the time-varying CDF $G(t)$. Instead, we assume that $G(t)$ remains  fixed during a one-hour time window and run a statistical test for independence on service times observed during consecutive one-hour windows. We note that the service time realizations are not assumed to be less than one hour, and we aggregate them only based on their starting time.

 We use Pearson's chi-square  test \cite{degroot2012probability} to check for independence of $S_i$ and $S_{i+1}$. We consider data for two months between March 1, 2018, and May 1, 2018. We bin the observed service times into five bins with equal (expected) mass.\footnote{The standard chi-square test assumes that data takes discrete values. So we  quantize the observed continuous data. We use $5$ quantization levels  and consider on average $\sim \hspace*{-2pt}100$ data points in each one-hour window.} In the chi-squared test we construct a contingency tables for the pairs of observations for consecutive service times. We compare the observed frequency for every possible pair values and the expected values when the consecutive service times are independent (null hypothesis). Under the independence assumption, the sum of squared differences between the observed frequencies and the expected values must follow a chi-square distribution. 
 
 Figure \ref{fig:assump-independence} shows the box plot of p-values of the chi-square test for $1464$ one-hour windows for every parking  location.\footnote{We only include the one-hour window if every bin has a frequency of at least five observations.} Given the consistent high values of the p-value for all parking locations, we cannot reject Assumption \ref{assump-independent} on service time independence, and thus, we assume that it holds for the observed data.   

%\red{Please explain this test}

\begin{figure}[t!]
	\begin{minipage}{0.48\textwidth}
		\centering
		\hspace*{-20pt}	\includegraphics[width=1.15\textwidth]{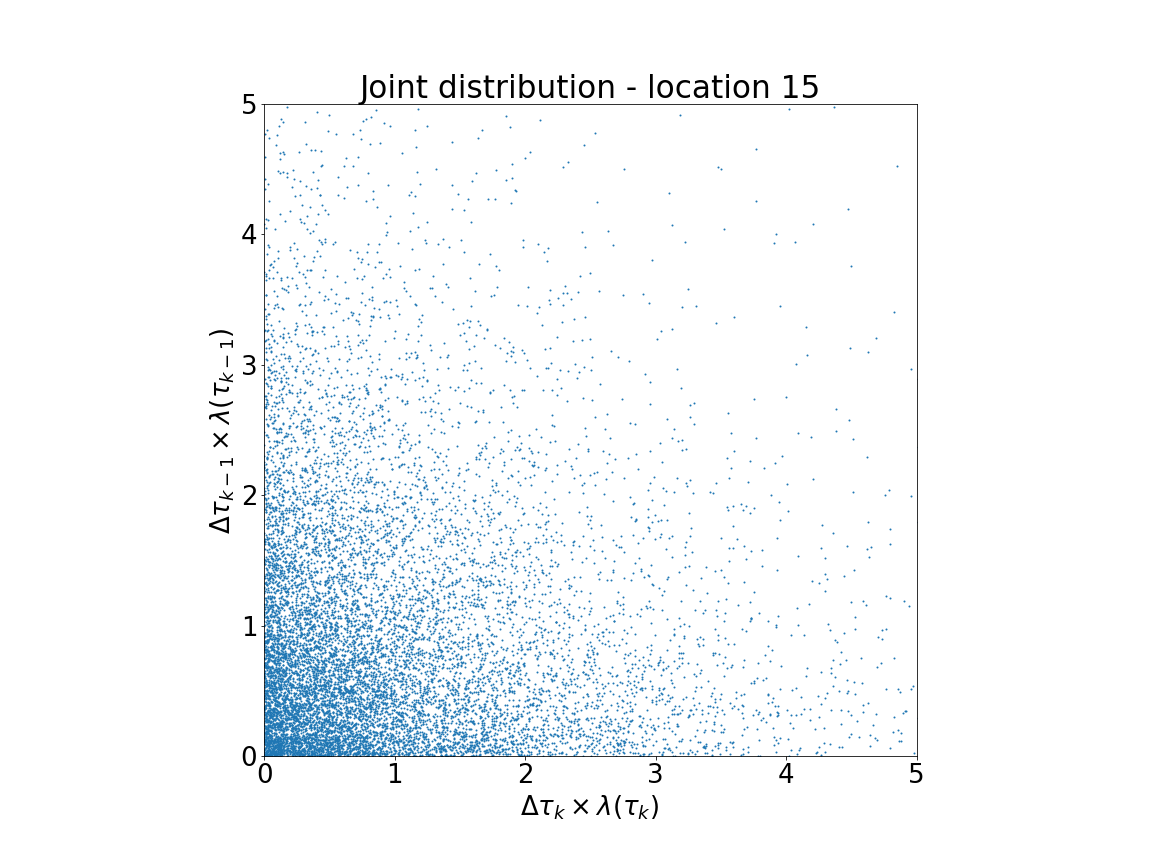}
		\caption{The scatter plot of consecutive normalized inter-arrival times for parking locations $15$ - $\sim 12500$ data points during a 10-day window from March $1$, to March $10$, 2018}
		\label{fig:assump-poisson-loc15}
	\end{minipage}%
	\hspace*{10pt}
	\begin{minipage}{0.48\textwidth}
		\centering
		\hspace*{-20pt}\includegraphics[width=1.15\textwidth]{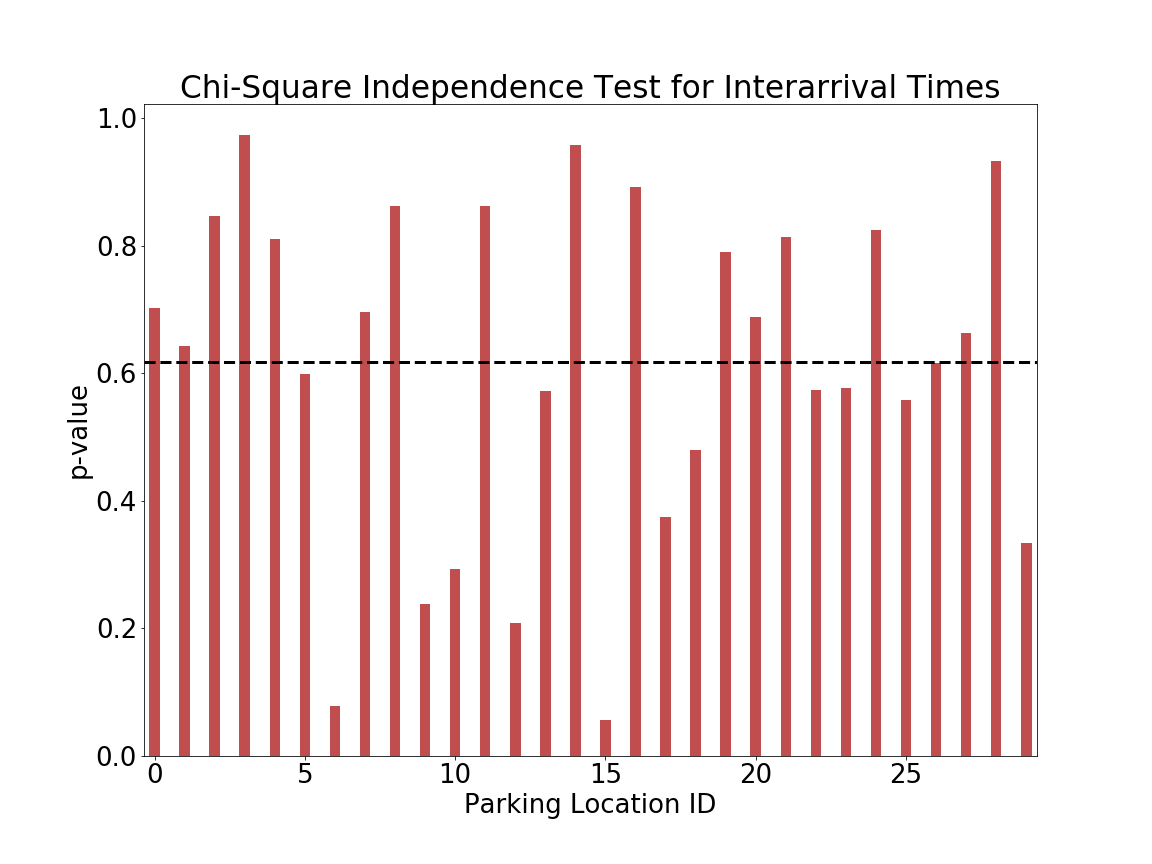}
		\caption{Pearson's chi-square independence test for normalized inter-arrival times  - The p-value for the 10-day window from March $1$, to March $10$, $2018$ for every parking location. The dashed line depicts the average p-value across all locations.}  
		\label{fig:assump-poisson-independence}
	\end{minipage}
\end{figure}%

\subsection{Poisson Arrival Process}

The arrival process in the $M(t)/G(t)/\infty$ queuing model is a non-homogeneous Poisson process. Therefore, we cannot directly apply a statistical test on the observed data to validate Assumption \ref{assump-poisson}. Similar to the approach taken in Section \ref{sec:-verification-independence}, we assume that the arrival rate $M(t)$ remains (approximately) constant during  one-hour windows; that is, during each one-hour window the arrival process is a homogeneous Poisson process. 

We say an arrival process is a homogeneous Poisson process if and only if 1) the inter-arrival times are independent, and 2) the inter-arrival times are exponentially distributed with mean $\frac{1}{M(t_k)}$ where $M(t_k)$ is the average arrival rate during the $k^{th}$ one-hour window.   
Let $\Delta \tau_i:=\tau_{i+1}-\tau_i$ denote the $i^{\text{th}}$ inter-arrival time, i.e. the time difference between the arrival time of vehicles $i$ and $i+1$. Define $\bar{\Delta}\tau_i :=(\tau_{i+1}-\tau_i)M(t_k)$ as the \textit{normalized} $i^{th}$ inter-arrival time where we assume that vehicle $i$ arrives during the $k^{th}$ one-hour window. We can show that the arrival process is a non-homogeneous Poisson process if and only if $\bar{\Delta} \tau_i$, $i\in\mathcal{N}$ are (approximately)\footnote{Note that we are making two approximations: 1) we assume that the arrival rate $M(t)$ is constant during each one-hour window, and 2) in computing $\bar{\Delta\tau}_i$, we neglect the possibility that trucks $i$ and $i+1$ arrive in two consecutive one-hour windows with $M(t_k)\neq M(t_{k+1})$.} i.i.d exponential random variables with mean $1$. 

The definition  of normalized inter-arrival times $\bar{\Delta}\tau_i$, $i\in\mathcal{N}$, enables us to validate Assumption \ref{assump-poisson} by pooling the observed data over a longer time horizon rather than considering each one-hour window separately. We calculate $\bar{\Delta}\tau_i$ by approximating $M(t_k)$ as the number of observed arrivals during the $k^{th}$ one-hour window. Figure \ref{fig:assump-poisson-loc15} shows the scatter plot of the consecutive normalized inter-arrival times for parking location $15$ during a 10-day window from March $1$, $2018$, to March $10$, 2018. We apply the Pearson's chi-square independence test for all parking locations over the same 10-day window. We bin the observed normalized inter-arrival times into eight bins with equal (expected) mass and follow a procedure for the chi-square test similar to the one described in Section \ref{sec:-verification-independence}. 

Figure \ref{fig:assump-poisson-independence} shows the result for every parking location. The average p-value (dashed line) is $0.62$ with the lowest p-values of $0.06$ for parking location $15$ and the highest p-value of $0.97$ for parking location $4$. Considering the (approximate) uniform spread of p-values between $[0,1]$ (which must happen under the null hypothesis \cite{degroot2012probability}) and the average p-value of $0.62$, we do not have any evidence against the independence of the normalized inter-arrival times, and thus, we can empirically verify it.   

%\red{Explain the chi-squared test in this context.}

\begin{figure}[t!]
	\begin{minipage}{0.48\textwidth}
		\centering
		\hspace*{-20pt}\includegraphics[width=1.15\textwidth]{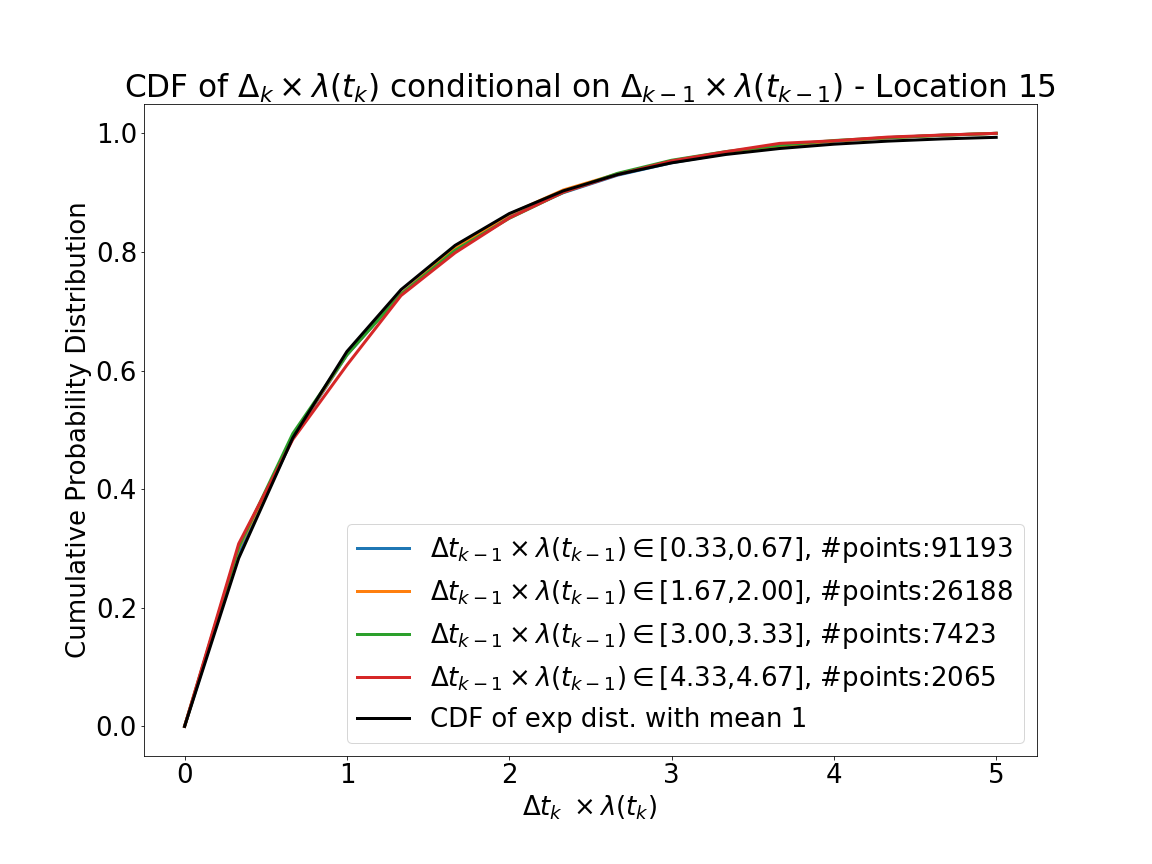}
		\caption{Probability distribution of $\bar{\Delta}\tau_k$ conditioned on $\bar{\Delta}\tau_{k-1}$ during the one-year window of $2017$ for parking location $15$.}
		\label{fig:assump-poisson-concdc}
	\end{minipage}%
	\hspace*{10pt}
	\begin{minipage}{0.48\textwidth}
		\centering
		\hspace*{-20pt}\includegraphics[width=1.15\textwidth]{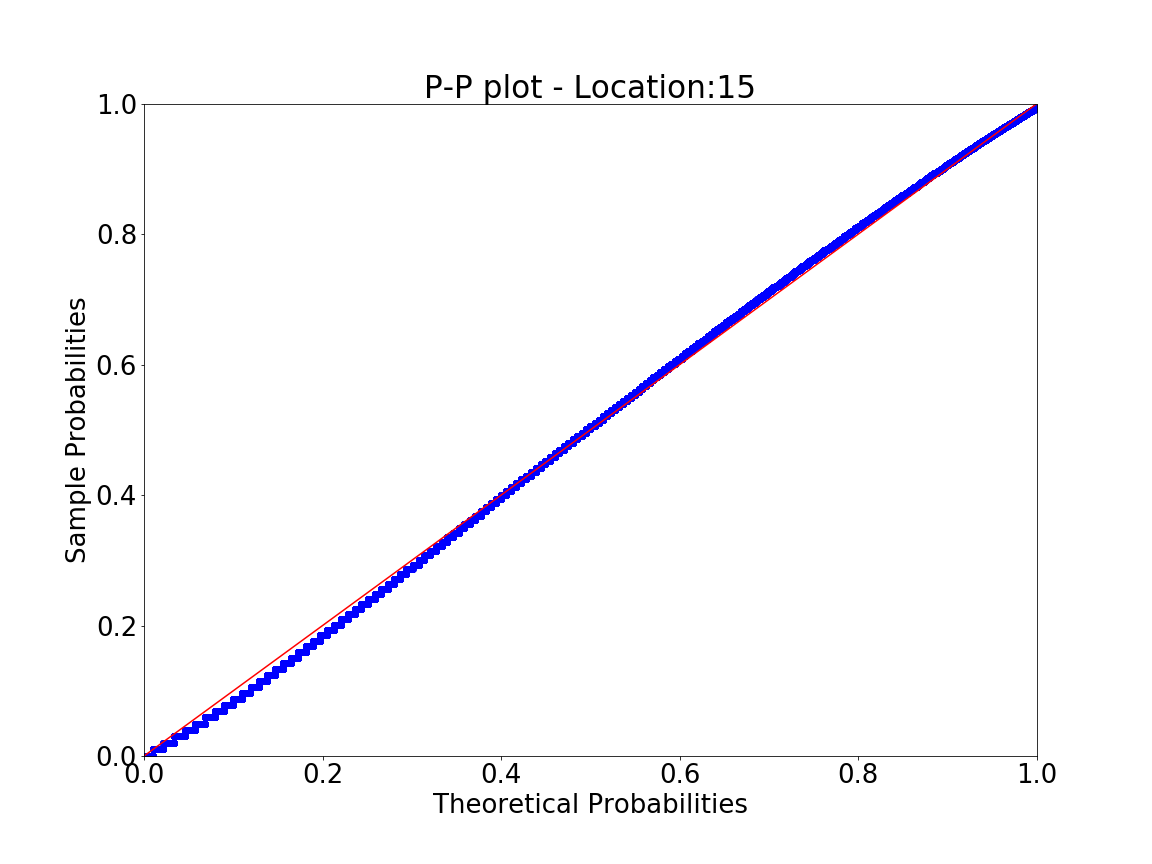}
		\caption{P-P plot of inter-arrival distribution distribution of $\bar{\Delta}\tau_k$ conditioned on $\bar{\Delta}\tau_{k-1}$ during the one-year window of $2017$ for parking location $15$.}
		\label{fig:assump-poisson-PP}
	\end{minipage}
\end{figure}%

\begin{figure}[t!]
	\begin{minipage}{0.48\textwidth}
		\centering
		\hspace*{-20pt}\includegraphics[width=1.15\textwidth]{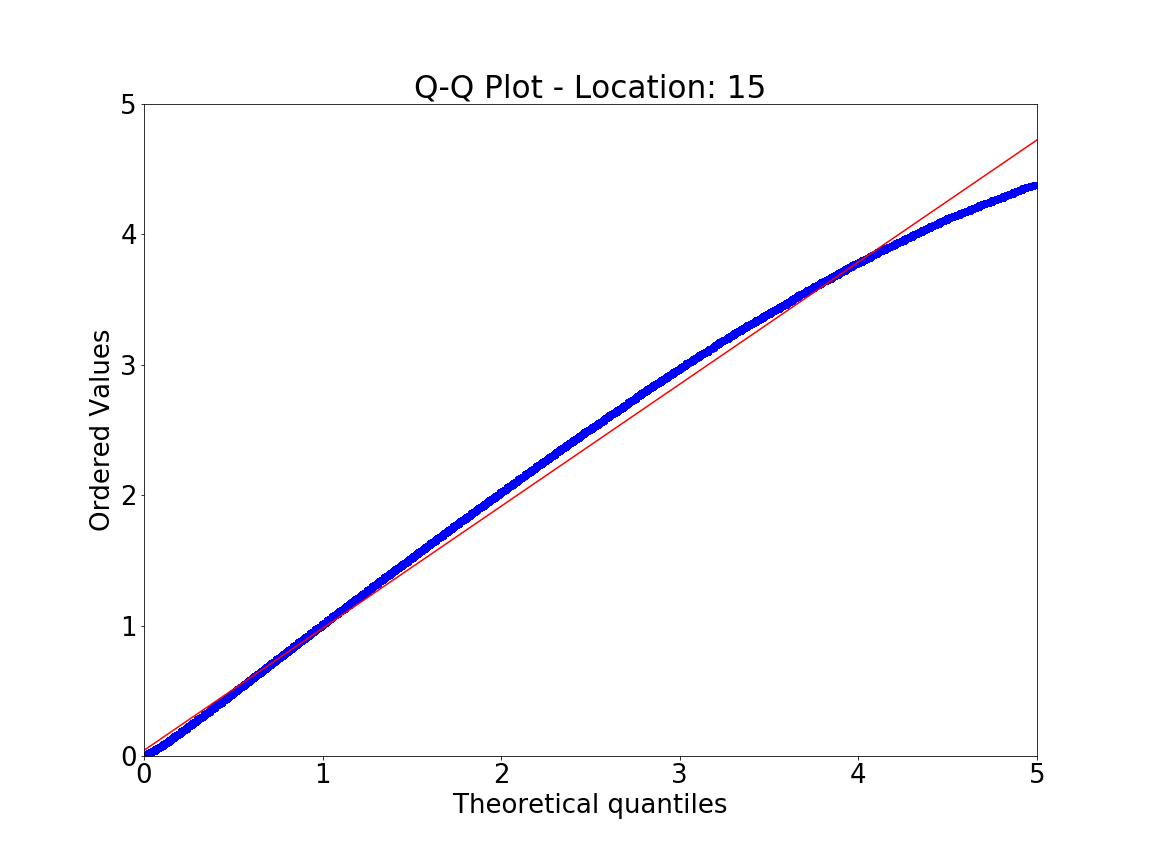}
		\caption{Q-Q plot of inter-arrival distribution distribution of $\bar{\Delta}\tau_k$ conditioned on $\bar{\Delta}\tau_{k-1}$ during the one-year window of $2017$ for parking location $15$.\vspace{22pt}}
		\label{fig:assump-poisson-QQ}
	\end{minipage}%
	\hspace*{10pt}
	\begin{minipage}{0.48\textwidth}
		\centering
		\hspace*{-20pt}\includegraphics[width=1.15\textwidth]{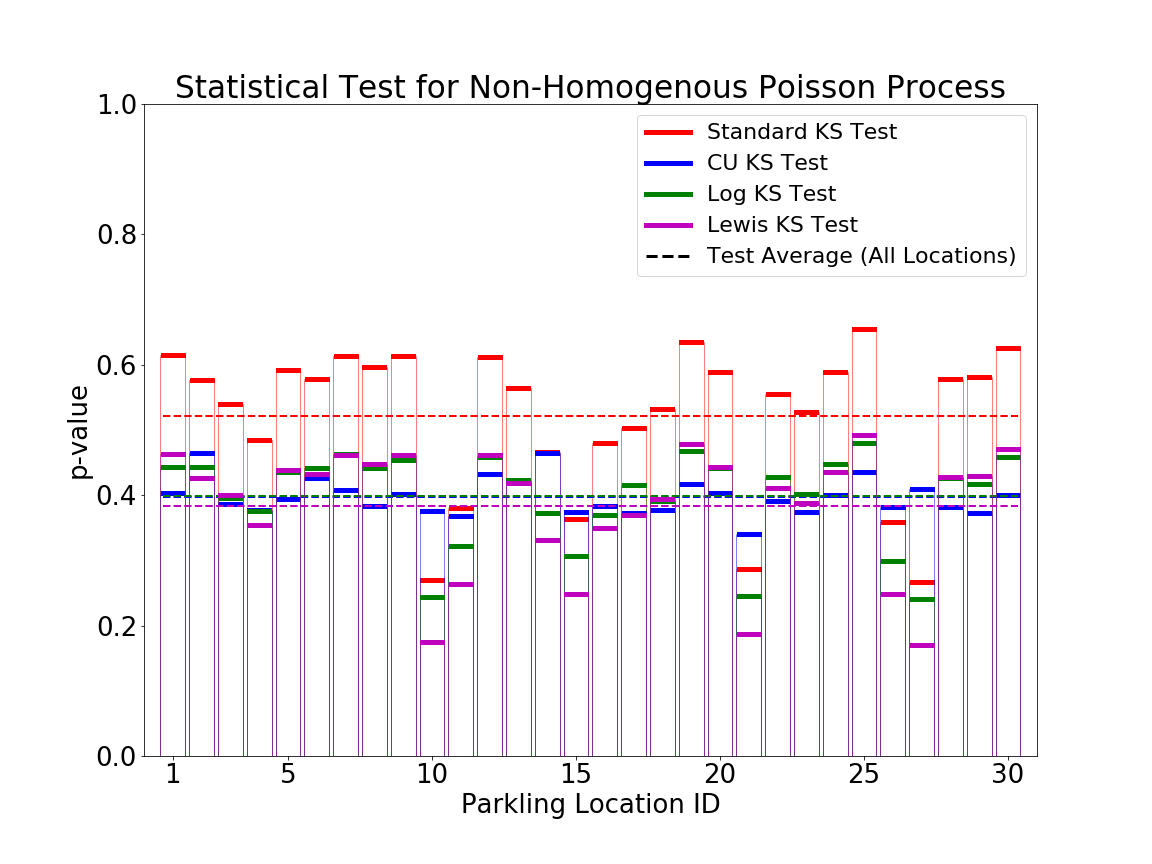}
		\caption{The average p-values for observed arrivals between January 1, 2018 to March 1, 2018: Standard KS test (red), CU KS test (blue), Log KS test (green), and Lewis KS test (purple). The dashed line present the average p-value across all locations.}
		\label{fig:assump-poisson-test}
	\end{minipage}
\end{figure}

Next, we compare the distribution of normalized inter-arrival times with that of an exponential distribution with mean $1$. We use the observed data during the one-year window of $2017$ with $\sim500,000$ observed arrivals. Figure \ref{fig:assump-poisson-concdc} shows the probability distribution of normalized inter-arrival times $\bar{\Delta}\tau_k$ conditioned on the value of $\bar{\Delta} \tau_{k-1}$ for location $15$. As can be noted in Figure \ref{fig:assump-poisson-concdc}, all conditional probability distributions closely follow the probability distribution of an exponential random variable with mean $1$. Figures \ref{fig:assump-poisson-PP} and \ref{fig:assump-poisson-QQ} depict the P-P plot and Q-Q plot of the probability distribution of inter-arrival time $\bar{\Delta}\tau_k$ versus the exponential probability distribution with mean $1$ for parking location $15$. According to the P-P and Q-Q plots, we can empirically verify the agreement between the observed data and the exponential probability distribution with mean $1$; %\footnote{We note that the two distributions are slightly mismatched for values greater than $4$ (that corresponds to tail probability of $e^{-4}\sim\%1$). Such a mismatch could point toward a substantial difference between the two distribution; nevertheless, considering the small probability of values greater than $4$ and the approximations we made in the calculation of $\bar{\Delta\tau_k}$, such a result seems satisfactory.} 
A similar analysis is preformed on all observed data to verify Assumption \ref{assump-poisson} for every parking location. 

\begin{remark}\label{remark:mismatch}
	Comparison of Figures \ref{fig:assump-poisson-PP} and \ref{fig:assump-poisson-QQ} reveals a slight mismatch between the empirical  and theoretical distributions. Specifically, compared to the theoretical distribution, the empirical distribution has a slightly higher mass over intermediate values (\textit{i.e.} higher slope in Figure \ref{fig:assump-poisson-PP} around $0.5$ probability) and a slightly lower mass over the extreme values (\textit{i.e.} lower slop in Figure \ref{fig:assump-poisson-PP} around $0$ and $1$ probability). While such a discrepancy may hint toward a slight mismatch between the model and the data, we argue that such a mismatch is, at least partially, due to the estimation process of the normalized inter-arrival times. To calculate the normalized inter-arrival times $\bar{\Delta} \tau_k$, we approximate the arrival rate during each one-hour window with the  total number of arrivals observed during that window. However, the total number of arrivals during each window is a random variable that is negatively correlated with the (average) observed inter-arrival times during that window; the higher the number of arrivals the lower the (average) inter-arrival times. Therefore, compared to the theoretical distribution, we expect the empirical distribution to have a slightly higher mass over intermediate values and a slightly lower mass over very low or high values; see  Figure \ref{fig:assump-poisson-PP}.        
\end{remark}

Non-homogeneous Poisson processes have also been used to model call centers \cite{ibrahim2012modeling,kim2014choosing}. In this literature, there are, in general, four types of statistical tests that have been used to empirically verify the non-homogeneous Poisson process component of the model. 
%These four statistical tests are as follows. 
\begin{enumerate}[i)]
	\item \textbf{Standard KS test:} The standard Kolmogorov-Smirnov (KS) test that compares the empirical CDF of normalized inter-arrival times $\bar{\Delta} \tau_k$ with that of $exp(1)$. 
	\item \textbf{CU KS test:} Under the Poisson process assumption, the arrival times during a time interval $[0,t]$ with fixed arrival rate are distributed as an order statistics of i.i.d uniform random variables over $[0,t]$. The conditional-uniform (CU) KS evaluates this hypothesis through a KS test.
	\item \textbf{Log KS test:} The Log KS test performs a similar KS test over a logarithmic transformation of arrival times during a time interval.\footnote{An exact definition of Log KS test can be found in \cite{kim2014choosing}.}
	\item \textbf{Lewis KS Test:} The Lewis test is a modification of CU KS test that offers a higher power against arrival processes with non-exponential CDF for inter-arrival times; however, as noted in \cite{kim2014choosing}, the Lewis KS test has lower power against dependent inter-arrival times compared to CU test. 
\end{enumerate} 

In contrast to our analysis so far, these four statistical tests evaluate whether both the independence and exponential distribution of inter-arrival times hold simultaneously, and thus, they do not provide a separate validation for each of them separately; Kim and Whitt \cite{kim2014choosing} provide an excellent survey of these test and discuss the strength and drawbacks of each one. 

We run these tests on our dataset to provide further evidence for the validity of Assumption \ref{assump-poisson}.
Figure \ref{fig:assump-poisson-test} shows the result of these four tests for each parking location. We use the observed data for a 60-day window from January $1$, $2018$, to March $1$, 2018.\footnote{As we discussed in Remark \ref{remark:mismatch}, there is a slight mismatch between the empirical distribution and the theoretical distribution for $exp(1)$. Therefore, we do not pool together all the data over the time interval of the two months to perform the test. Otherwise, the KS test strongly rejects the null hypothesis because of such persistent slight mismatch between the two distribution for large sample sizes; see \cite{ibrahim2013forecasting} for a survey of challenges in predicting a doubly stochastic Poisson Process. Alternatively, we apply the tests separately over each one-hour window, with an average of approximately $50$ data points, and report the average value over these two months.} Given the p-values calculated across all parking locations for each test in Figure \ref{fig:assump-poisson-test}, we cannot reject Assumption \ref{assump-poisson}.

\begin{figure}[t!]
	\begin{minipage}{0.48\textwidth}
		\centering
		\hspace*{-20pt}\includegraphics[width=1.15\textwidth]{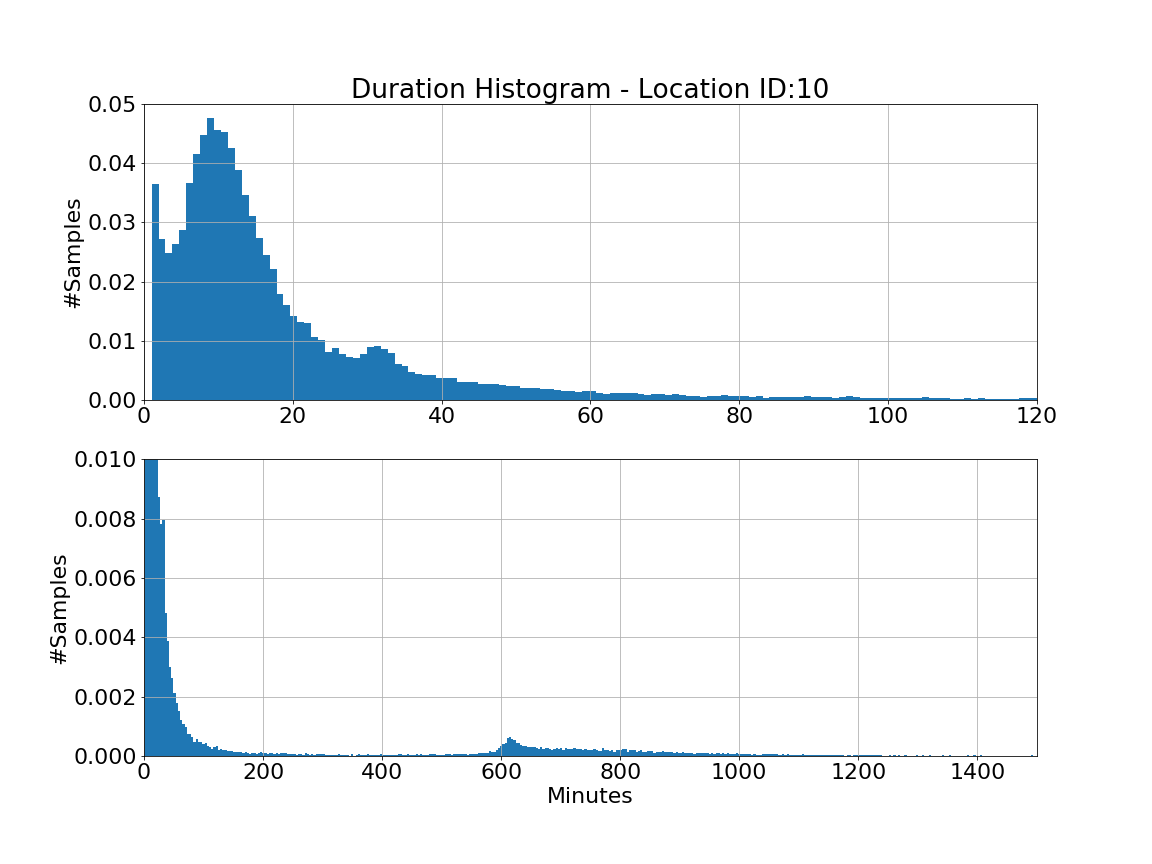}
		\caption{The empirical distribution of service time for parking location $10$: (top plot) empirical distribution over interval $[0,120]$ minutes and (bottom plot) empirical distribution over interval $[0,1500]$ minutes.}
		\label{fig:servicedist-loc10}
	\end{minipage}%
	\hspace*{10pt}
	\begin{minipage}{0.48\textwidth}
		\centering
		\vspace*{10pt}
		\hspace*{-20pt}\includegraphics[width=1.15\textwidth]{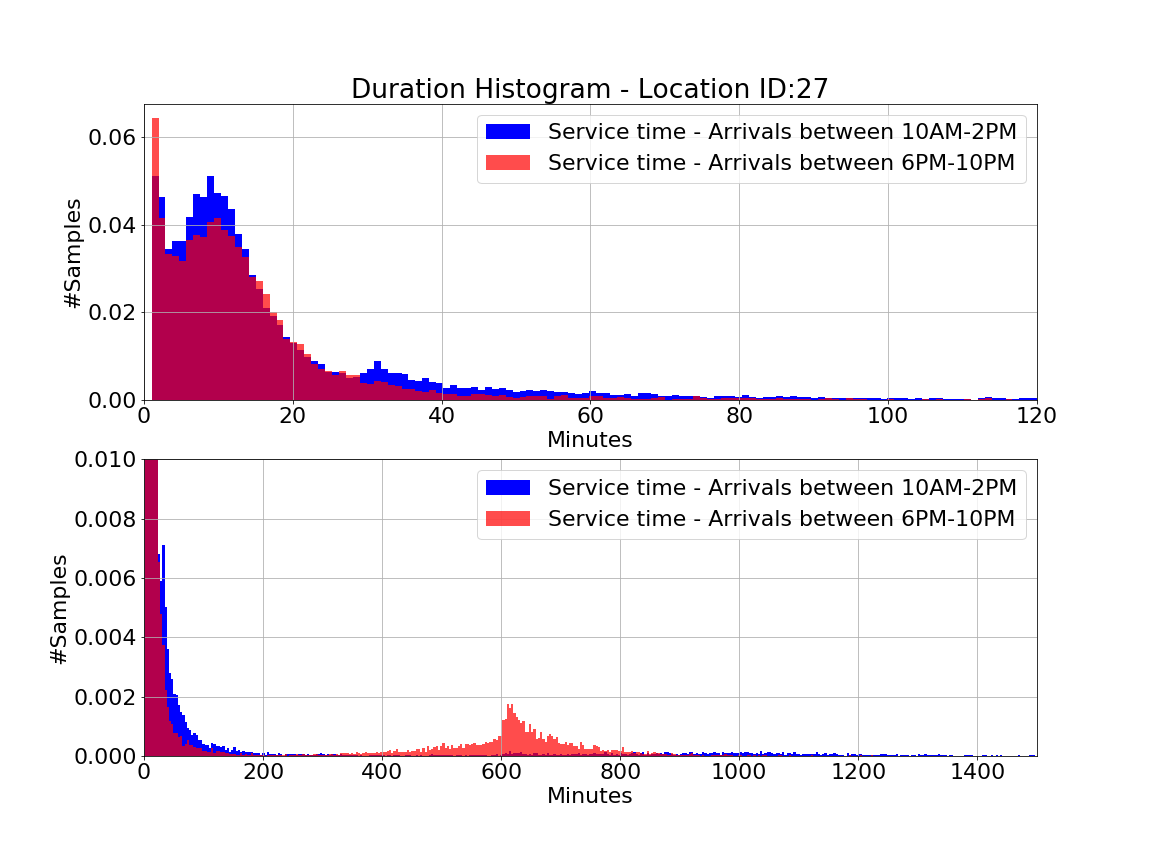}
		\caption{The time variation of the empirical distribution of service time for parking location $27$: (top plot) empirical distribution over interval $[0,120]$ minutes and (bottom plot) empirical distribution over interval $[0,1500]$ minutes. }
		\label{fig:servicedist-time-loc27}
	\end{minipage}
\end{figure}%

\section{Analysis}\label{sec:analysis}

In the $M(t)/G(t)/\infty$ model, both the arrival process and the service time distribution are time-varying. In this section, we examine the time variation of these two components. We utilize our findings to expand on our basic queuing model to construct a prediction method in Section \ref{sec:prediction}.

\subsection{Service Times and Heterogeneous Populations}\label{sec:analysis-service}
The main challenge in characterizing  the $M(t)/G(t)/\infty$ queuing model is to capture the time variation in the service time distribution $G(t)$. This is because we need to model a time-varying functional $G(t)$ in contrast to a time-varying scalar for the arrival rate $M(t)$. A standard approach is to parametrize the distribution  $G(t)$, and reduce the time variation in the functional space to the time variation in a (finite) vector space. In the following, we propose such a parametrization of $G(t)$ based on the analysis of observed service times and our domain-specific knowledge.

 \begin{figure}[t!]
 	\begin{minipage}{0.48\textwidth}
 		\centering
 		\hspace*{-20pt}\includegraphics[width=1.15\textwidth]{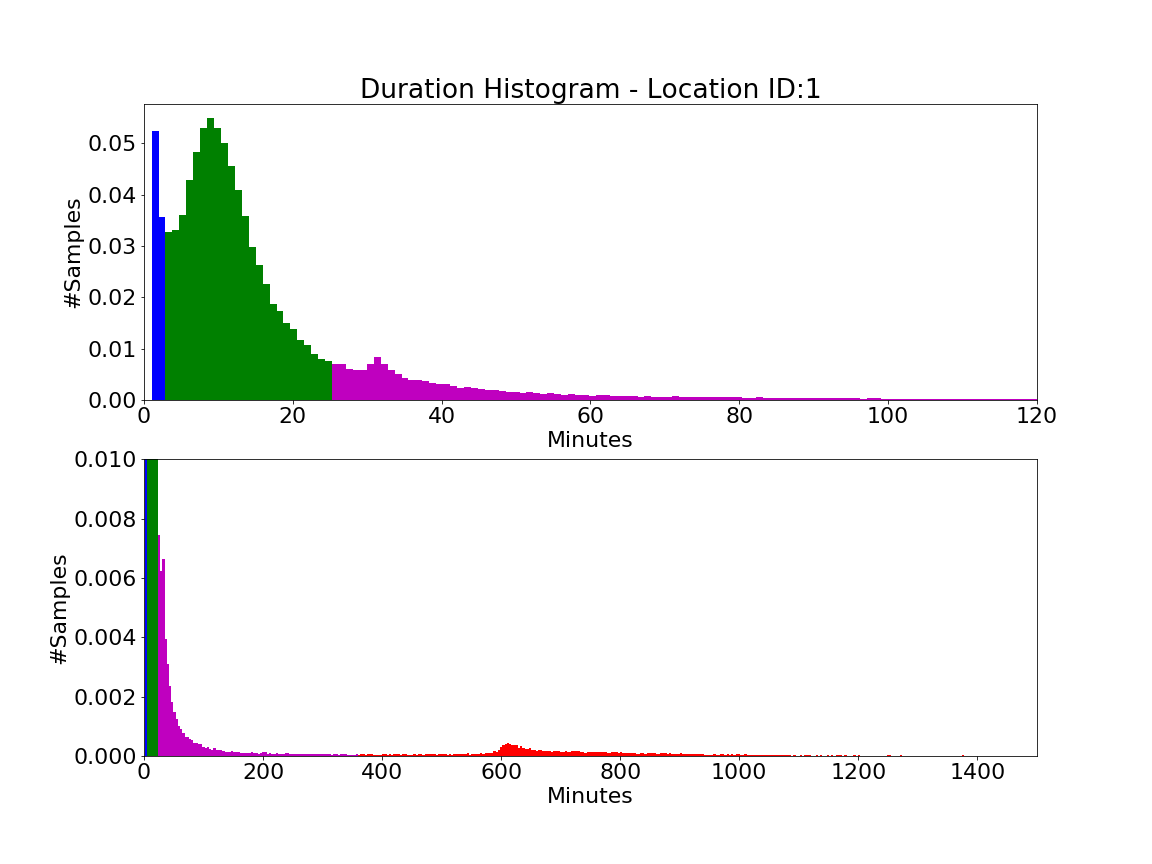}
 		\caption{The time variation of the empirical distribution of service time for parking location $1$: time variation over interval $[0,120]$ minutes (top plot) and time variation over interval $[0,1500]$ minutes (bottom plot).}
 		\label{fig:servicedist-pop-loc10}
 		%\label{fig:servicedist-loc10}
 	\end{minipage}%
 	\hspace*{10pt}
 	\begin{minipage}{0.48\textwidth}
 		\centering
 		\hspace*{-20pt}\includegraphics[width=1.15\textwidth]{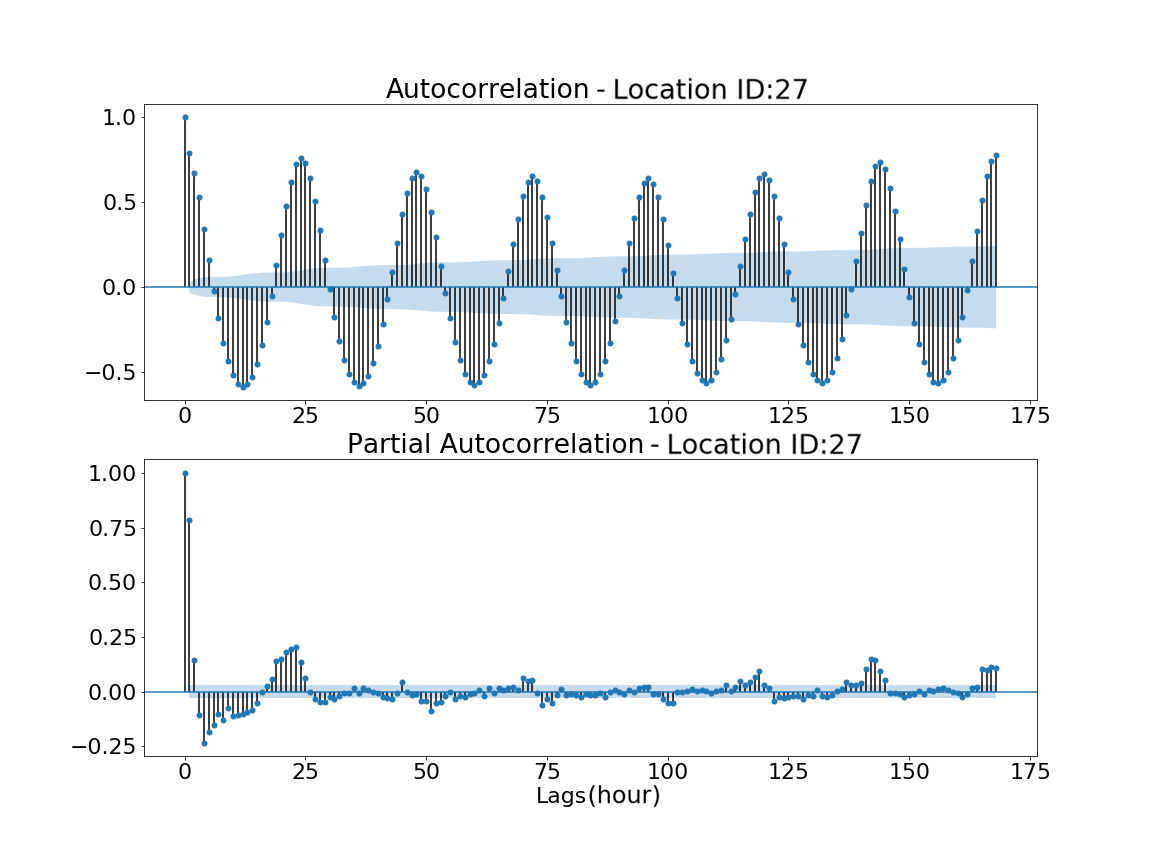}
 		\caption{The ACF (top plot) andPACF (bottom plot) for aggregate arrival process for location $27$. }
 		\label{fig:CF-27}
 	\end{minipage}
 \end{figure}%
 
 We start by examining the empirical distribution of the service times observed over a time interval. Figure \ref{fig:servicedist-loc10} shows the empirical service time distribution $G(t)$ based on the observed data during the span of four months between January 1, 2018, to May 1, 2018 for parking location $10$; we omitted service times of less than one minute to not distort the scale for the rest of the distribution. Four distinct peaks/modes in the empirical distribution can be noticed in the following intervals: 1) times between $0$ to $5$ minutes, 2) times between $5$ to $\sim25$ minutes, 3) times between $25$ and $~\sim360$ minutes, and 4) times larger than $360$ minutes. A similar pattern can be observed across all locations though there are variations in the height of each peak. We underscore that the empirical service time distribution is quite different from an exponential random variable that is assumed in most of the literature on queuing models for parking. 

Next, we explore the variation of the service time distribution over time. Figure \ref{fig:servicedist-time-loc27}  shows the variation of service time distribution during the day for location $27$ during the same span of time used in Figure \ref{fig:servicedist-loc10} for location $10$. We note that most of the variation observed in the service time distribution $G(t)$ during the day can be captured by adjusting the weight of each peak/mode identified above. Consequently, we propose to decompose the time-varying $G(t)$ into four heterogeneous populations of arrivals, each with a \textit{time-independent} service time distribution $G_i$, $i,=1,2,3,4$, as
\begin{align}
G(t)=\lambda_1(t)G_1+\lambda_2(t)G_2+\lambda_3(t)G_3+\lambda_4(t)G_4,
\end{align} 
 where $\lambda_i(t)$ denotes the probability that an arriving trucks at $t$ has a service time distribution of $G_i$. Based on our observation, we choose distribution $G_i$, $i,=1,\cdots,4$, with \textit{non-overlapping} support and possible justifications as follows:
  \begin{enumerate}[i)]
 	\item \textbf{Very short stops ($0-5$ minutes):} A class of arrivals that stop to (possibly) check the vehicle, buy an item from store, \textit{etc}.
 	\item \textbf{Short stops ($5-25$ minutes):} A class of arrivals that stop to (possibly) get a quick bites, take a short break, short maintenance of vehicle, \textit{etc}.
 	\item \textbf{Normal stops ($25-360$ minutes):} A class of arrivals that stop to (possibly) get a long break, eat a meal, \textit{etc}.
 	\item \textbf{Long stops (longer than $6$ hours):} A class of arrivals that stop to sleep over night, waiting for vehicle repair, \textit{etc}.
 \end{enumerate}
  
 Figure \ref{fig:servicedist-pop-loc10} shows the above four classes of populations for parking location $1$. For each parking location, we define $G_i$ as the empirical service time distribution observed over its support; as a result, some aspects of location-specific variations in the empirical distribution $G(t)$ are captured through this definition of $G_i$.   We note that one can consider alternative approaches to parametrize $G(t)$; for instance, one can consider decomposing $G(t)$ into a mixture of Gamma distributions with overlapping support. However, our experimentation with such alternatives did not result in a consistent performance improvement. Moreover, our  choice of non-overlapping support offers a computational advantage for prediction purposes in both the training and evaluation phases, and reduces the number of parameters we need to identify for each parking location separately. We further discuss the potential advantage of developing a more sophisticated technique in Section \ref{sec:prediction}.

 \begin{figure}[t!]
 	\begin{minipage}{0.48\textwidth}
 		\centering
 		\hspace*{-20pt}\includegraphics[width=1.15\textwidth]{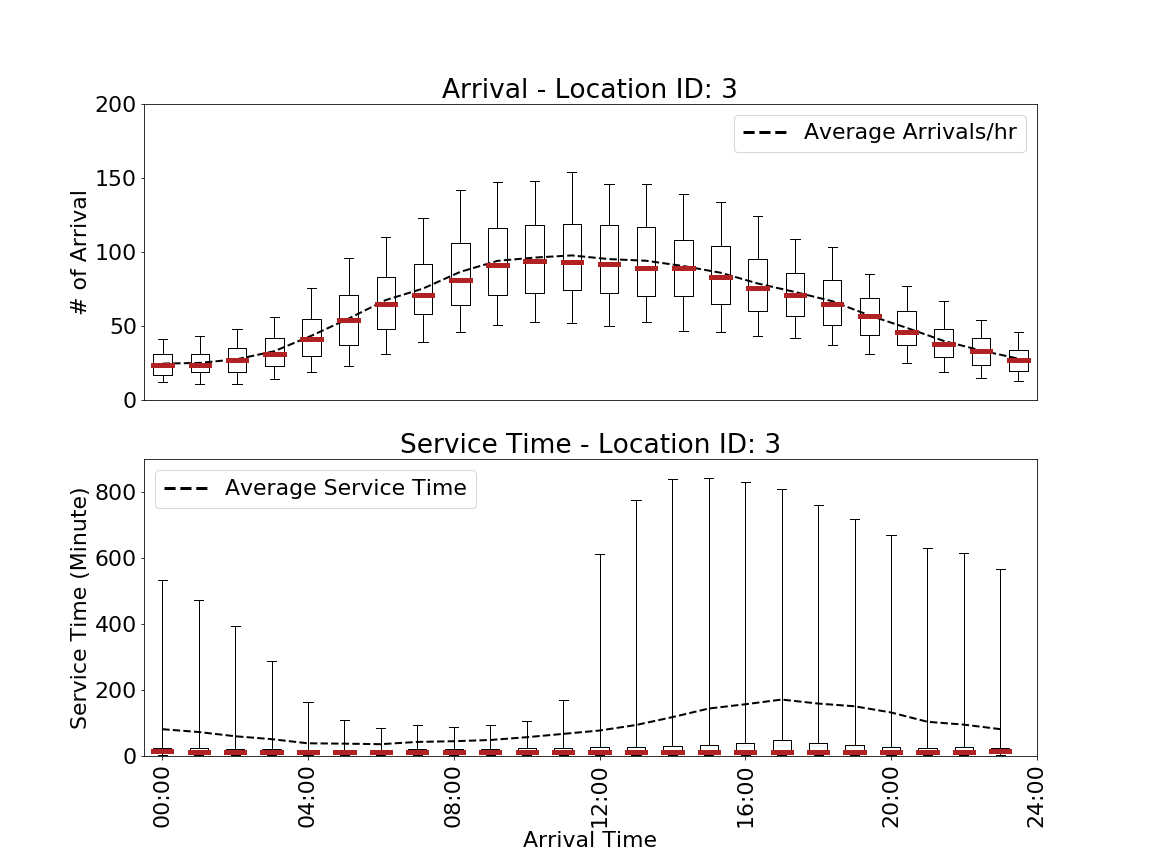}
 		\caption{Box plot of total arrivals  (top plot) and  box plot of service times during the day for location $3$.}
 		\label{fig:day-total-3}
 	\end{minipage}%
 	\hspace*{10pt}
 	\begin{minipage}{0.48\textwidth}
 		\centering
 		\hspace*{-20pt}\includegraphics[width=1.15\textwidth]{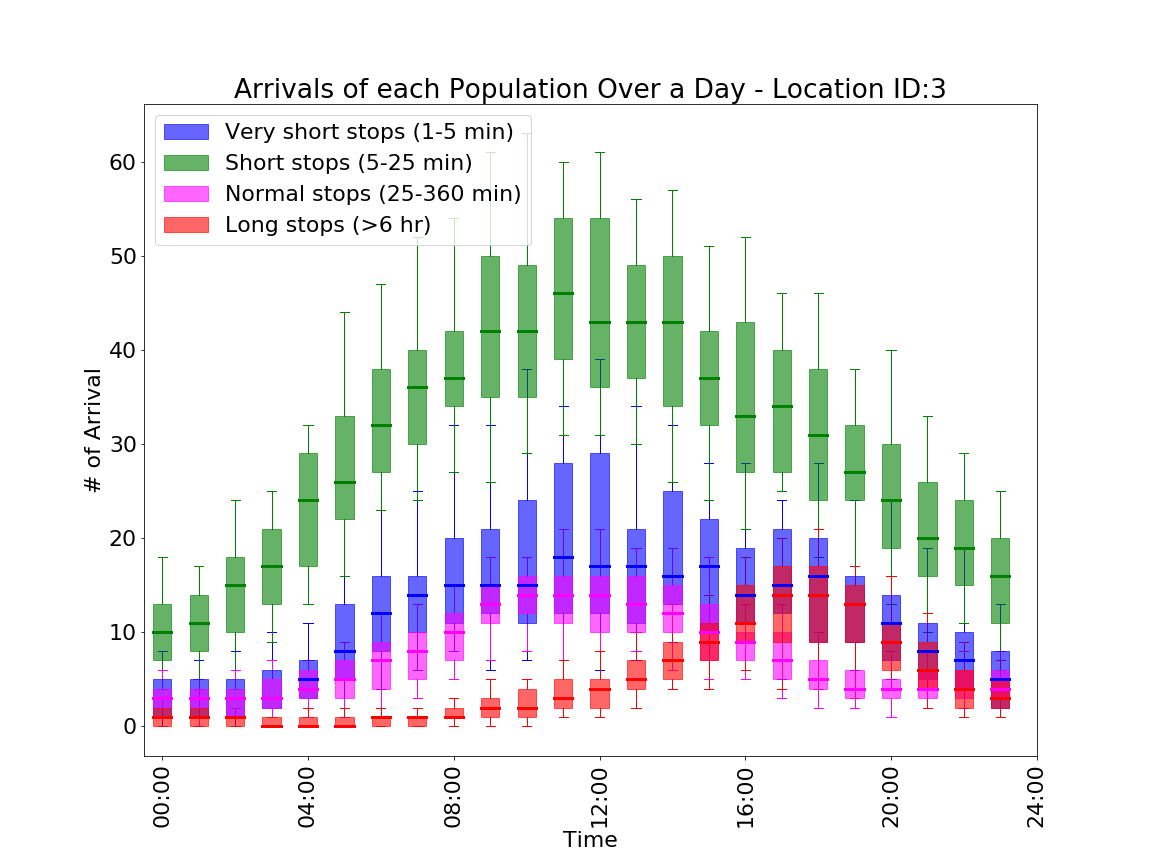}
 		\caption{Box plot of arrivals $M_i(t)$ for each Population during the day for location $3$.}
 		\label{fig:arrival-boxplot-pop-3}
 	\end{minipage}
 \end{figure}%

\subsection{Seasonal Patterns}\label{sec:analysis-arrival}

With the proposed decomposition of arrivals into four heterogeneous populations, to characterize the $M(t)/G(t)/\infty$ queuing model, we need to describe the arrival processes for each of these populations. In this section, we study the seasonal patterns present in the arrivals for these populations as well as the aggregate arrivals. We use these seasonal patterns to construct a predictive model for the arrival rates in Section \ref{sec:analysis-arima}.

Figure \ref{fig:CF-27} shows the autocorrelation function (ACF) and partial autocorrelation function (PACF) for the aggregate arrival process for parking location $27$; similar results hold for each individual population and for other parking lots. The ACF and PACF plots suggest the existence of seasonal patterns due to time of the day (inter-day) and weekday (intra-day) effects.

\textbf{Time-of-day effect:} Figure \ref{fig:day-total-3} shows the variation in total arrival rate $M(t)$ and service time during the day for parking location $3$ during the span of one year between May 1, 2017 to May 1, 2018. The number of arrivals increases during working hours, reaching its maximum around noon, and then decreases reaching its lowest value during mid-night. The average service time distribution is the lowest for arriving vehicles arriving early in the morning, and is highest for vehicles arriving in the evening, some of them probably staying overnight.     

\begin{figure}[t!]
	\begin{minipage}{0.48\textwidth}
		\centering
		\hspace*{-20pt}\includegraphics[width=1.15\textwidth]{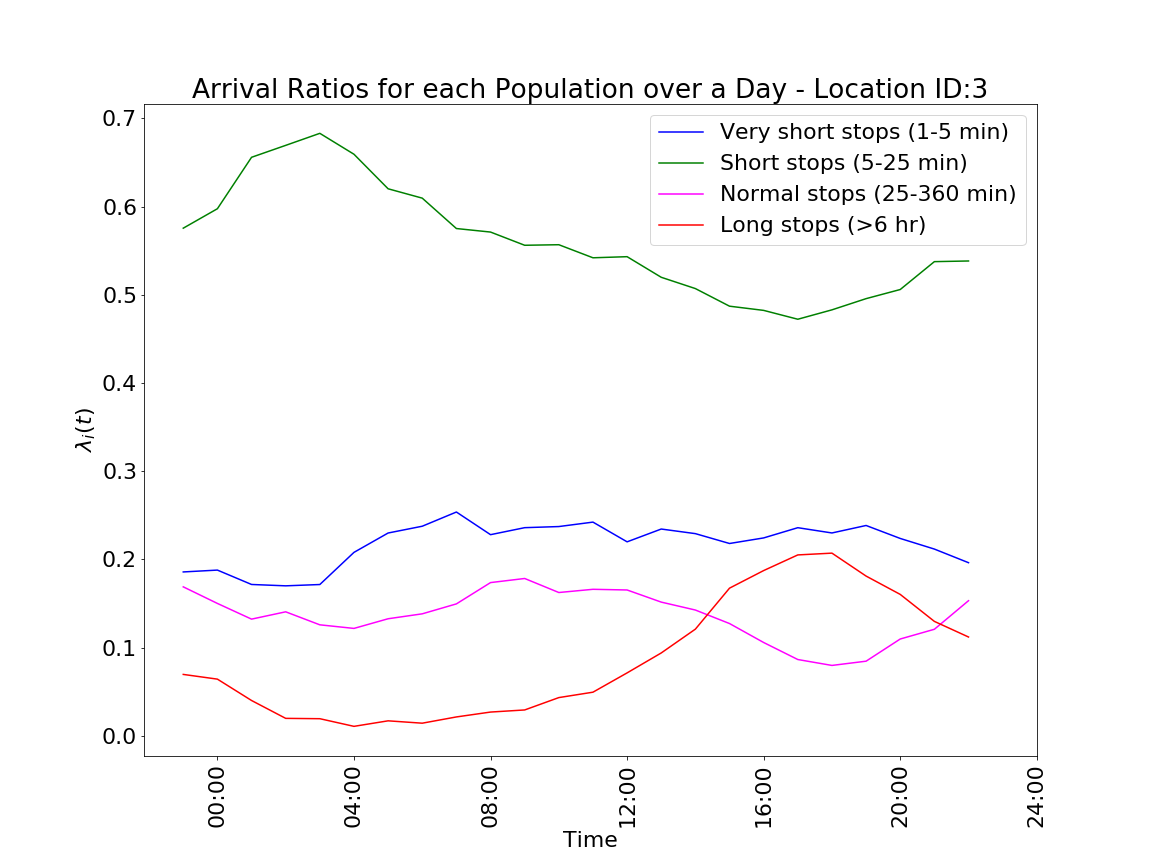}
		\caption{Share of each population $\lambda_i(t)$ during the day for location $3$.}
		\label{fig:arrival-ratio-pop-3}
	\end{minipage}%\vspace{-30pt}
	\hspace*{10pt}
	\begin{minipage}{0.48\textwidth}
		\centering
		\hspace*{-20pt}\includegraphics[width=1.15\textwidth]{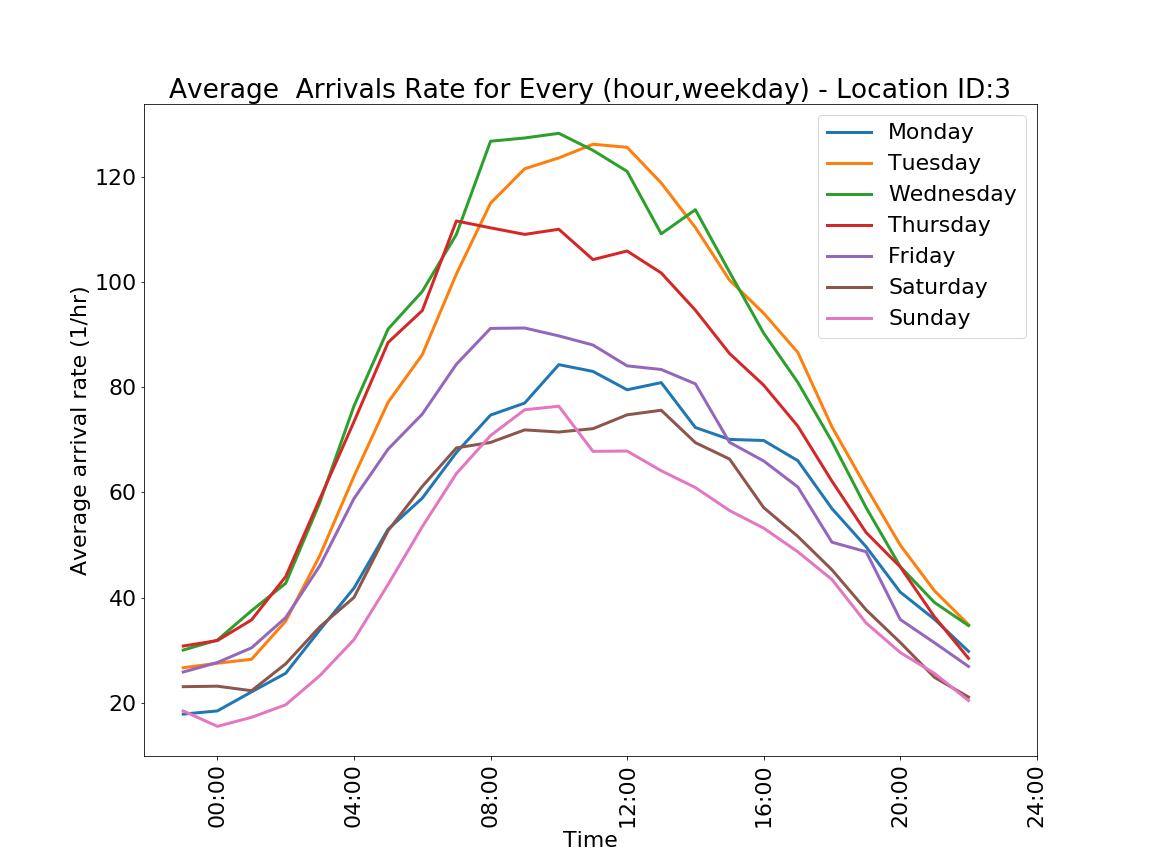}
		\caption{Average arrival rate $M(t)$ during the day for each weekday - parking location $3$.}
		\label{fig:arrival-weekday-3}
	\end{minipage}
\end{figure}

Define $\lambda_i(t):=\frac{M_i(t)}{M(t)}$ as the share of population $i$ from total arrival rates $M(t)$, where $M_i(t)$ denotes the number of arrivals (arrival rate) with service times that belongs to population $i$, $i=1,\ldots,4$. Figures \ref{fig:arrival-boxplot-pop-3} and \ref{fig:arrival-ratio-pop-3} show the variation of $M_i(t)$ and $\lambda_i(t)$ during the day. Consistent with Figure \ref{fig:day-total-3}, the number of arrivals for the first three populations increases during the day and decreases over night. By contrast, the number of arrivals for the fourth population, which has a longer service time, reaches its highest value during evenings.  %\blue{How are $M_i(t)$ calculated from data?}

\begin{figure}[t!]
		\begin{minipage}{0.48\textwidth}
			\centering
			\hspace*{-20pt}\includegraphics[width=1.15\textwidth]{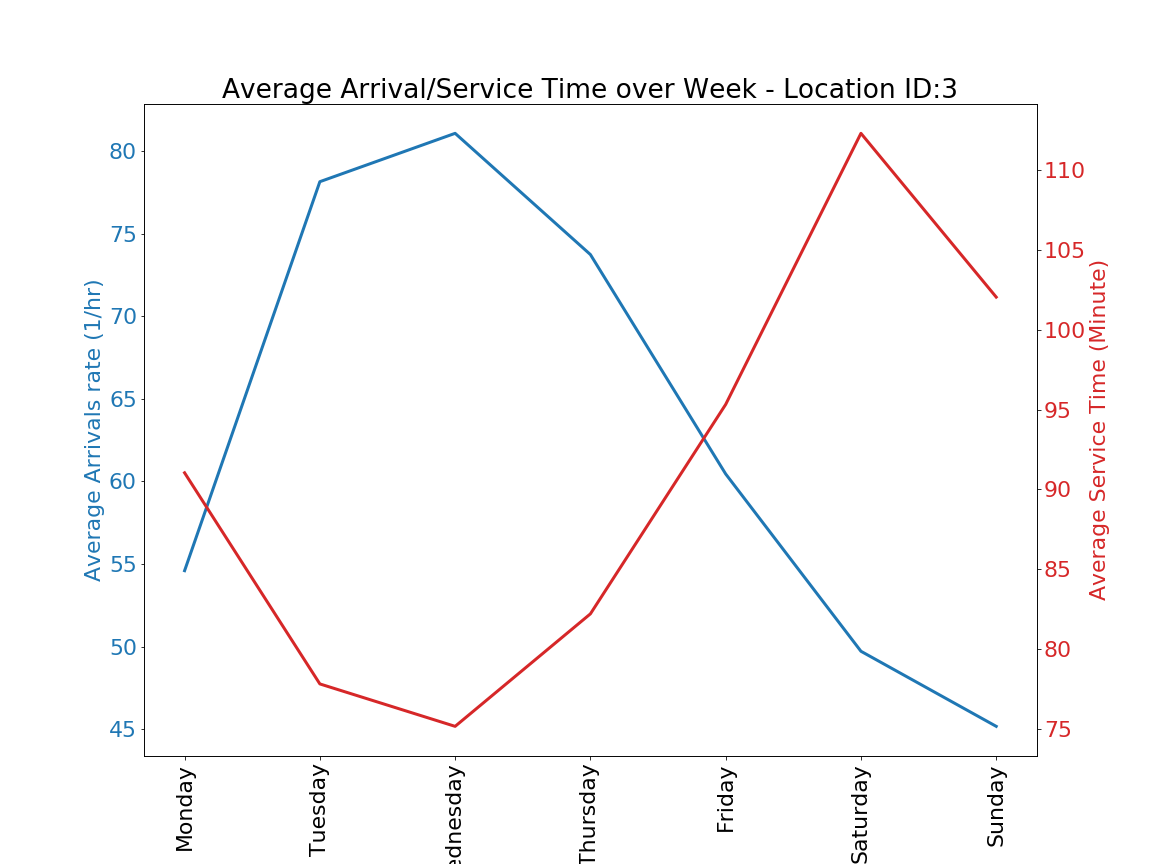}
			\caption{Average Arrival rate and service time for each weekday - parking location $3$}
			\label{fig:average-weekday-3}
		\end{minipage}	\hspace*{10pt}
	\begin{minipage}{0.48\textwidth}
		\centering
		\hspace*{-20pt}\includegraphics[width=1.15\textwidth]{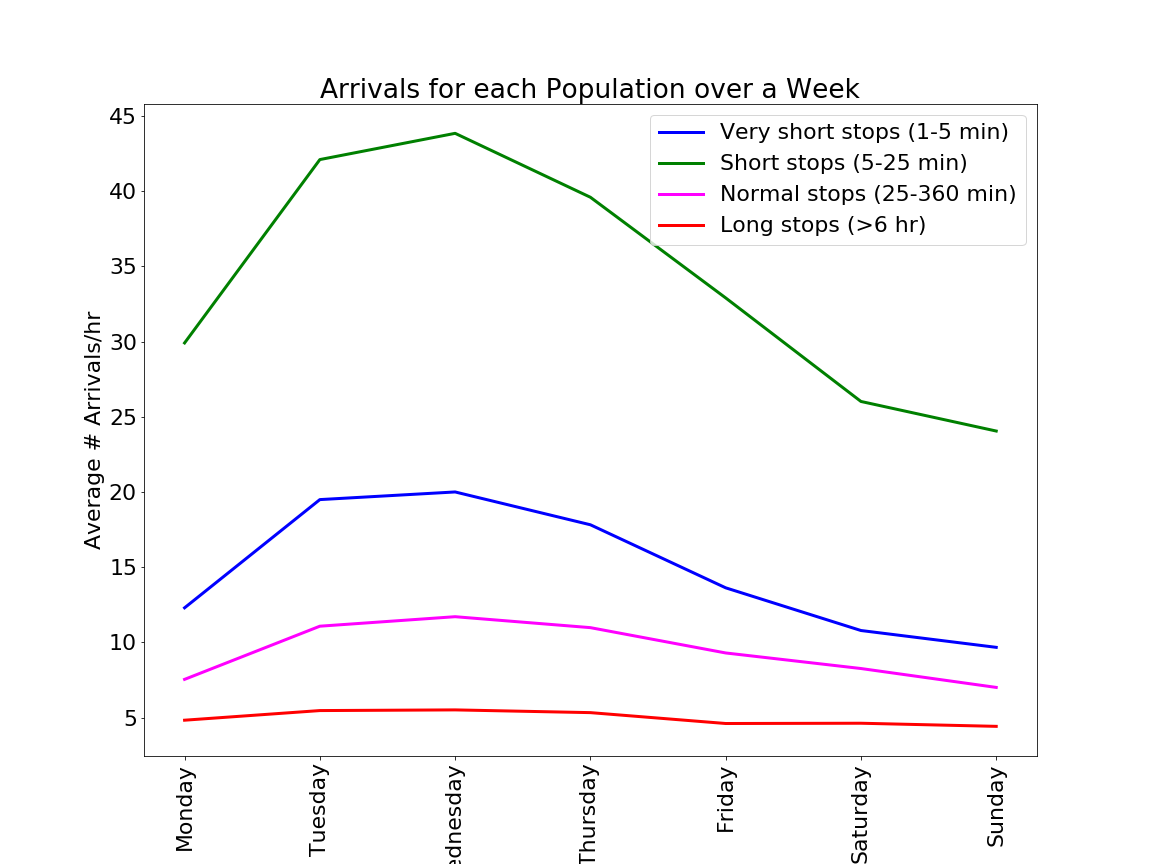}
		\caption{Average arrivals rates $M_i(t)$ for each population for every weekday - parking location $3$}
		\label{fig:arrival-pop-weekday-3}
	\end{minipage}%
%	\hspace*{10pt}
%	\begin{minipage}{0.48\textwidth}
%		\centering
%		\hspace*{-20pt}\includegraphics[width=1.15\textwidth]{Plots/arrival-pop-ratio-week3}
%		\caption{Average arrival ratio $\lambda_i(t)$ for each weekday - parking location $3$}
%		\label{fig:ratio-pop-weekday-3}
%	\end{minipage}
\end{figure}

\textbf{Day-of-week effect:} Figure \ref{fig:arrival-weekday-3} shows the variation of the average arrival rate during the day for each weekday. We note that the variation of arrivals during the day follows a similar pattern for all weekdays, with some days exhibiting a higher overall traffic compared to the others. Figure \ref{fig:average-weekday-3} depicts the variation of the day-average arrival rate and service time  
during the week. Consistent with our observation from Figure \ref{fig:arrival-weekday-3}, it appears that during Tuesday-Thursday we observe higher arrival rates while truck traffic is lower over the weekend. On the other hand,  the average service time is lower during Tuesday-Thursday and is higher over the weekend.    
Figure \ref{fig:arrival-pop-weekday-3} shows the average arrival rate $M_i(t)$ for each population over a week.

\textbf{Other effects:} We have also investigated for other possible seasonal patterns or trends such as monthly variation, holiday season, or the effect of sunrise and sunset times. However, no additional seasonal pattern or trend was identified in our analysis. We note that, however, the existing data with a span of 16 months does not offer enough observations to detect long-term patterns/trends. For the same reason, we also did not attempt to single out specific days during an annual calendar such as Thanksgiving day or Christmas day.    

\section{A Mixed-Effects Model for Arrivals}\label{sec:analysis-arima}

\begin{figure}[t!]
	\begin{minipage}{0.48\textwidth}
		\centering
		\hspace*{-20pt}\includegraphics[width=1.15\textwidth]{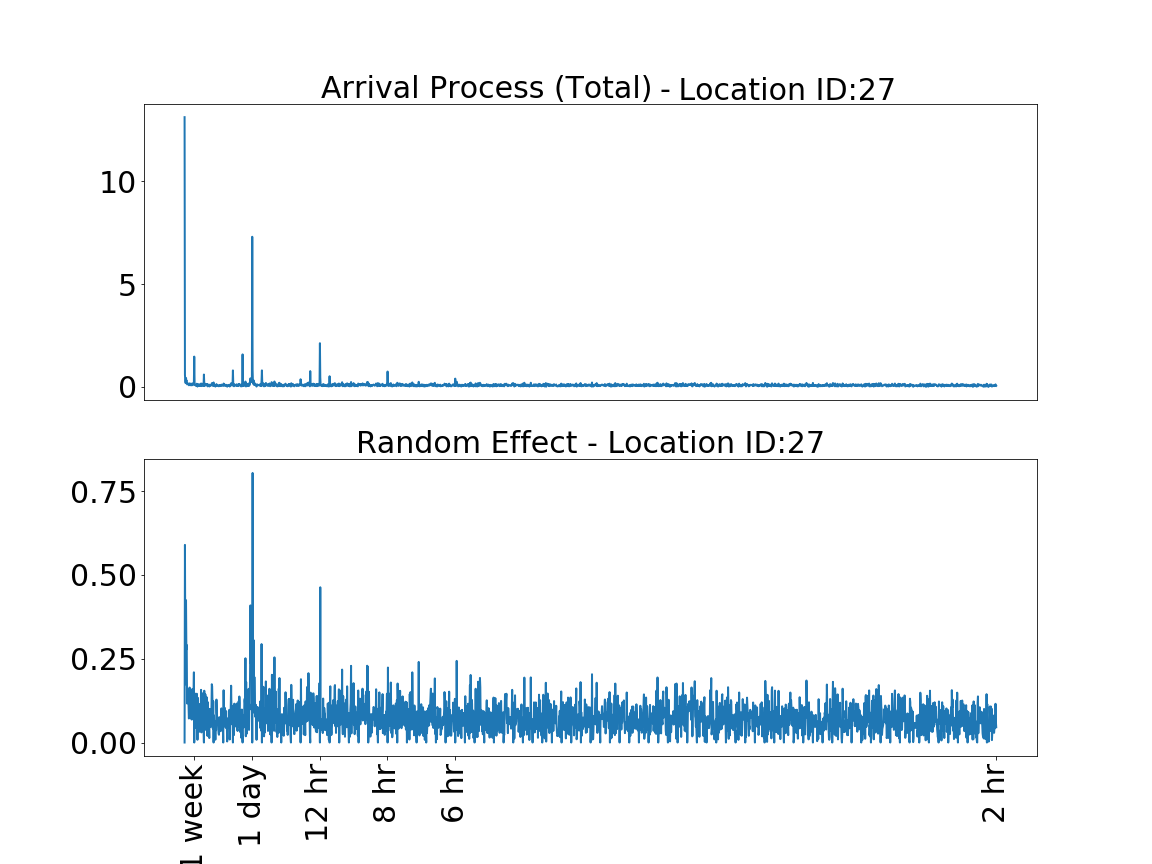}
		\vspace*{-17pt}
		\caption{ Fourier transform of arrival process $M(t)$ (top) and the residuals in additive model (bottom) for parking location $27$.  }
		\label{fig:FT-27}
	\end{minipage}%
	\hspace*{10pt}
	\begin{minipage}{0.48\textwidth}
		\centering
		\vspace*{9pt}
		\hspace*{-23pt}\includegraphics[width=1.15\textwidth]{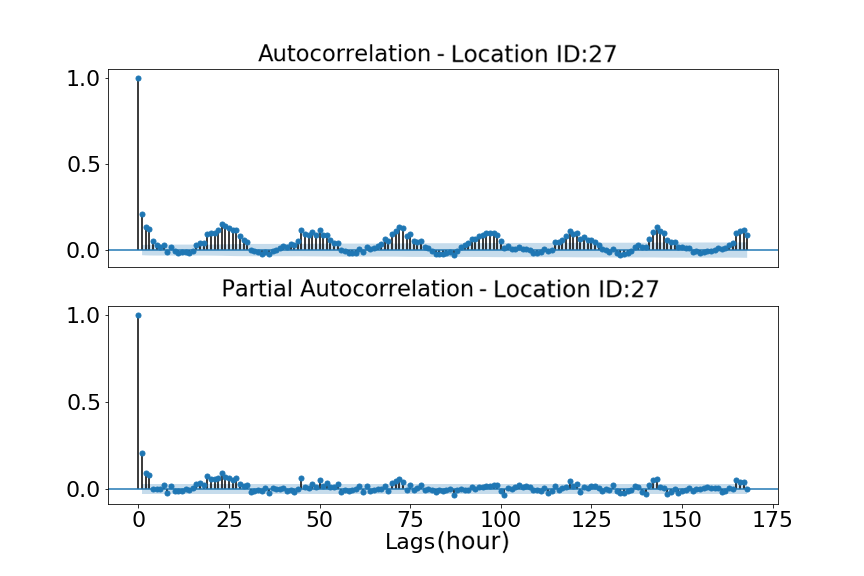}
				\vspace*{-6pt}
		\caption{The ACF (top) and PACF (bottom) for random effect $X_4(t)$ for population $4$ with long stops - location $27$}		
		\label{fig:CF-R-27}
	\end{minipage}
\end{figure}

Based on the analysis results of Section \ref{sec:analysis}, we consider a mixed effects model where the arrival rates at each time depend on the fixed seasonal effects we identified above, i.e. the time-of-day and day-of-week effects, and an additional random effect. Let $h(t)$ and $d(t)$ denote the time of the day and day of the week  corresponding to time $t$. Formally, for arrival process $M_i(t)$ for population $i=1,\cdots,4$, we assume that
\begin{align}\label{eq:mixedeffect}
	 M_i(t)=\sum_{j=0}^{6}\alpha_j\mathbbm{1}_{\{d(t)=j\}}+\sum_{i=0}^{23}\beta_k\mathbbm{1}_{\{h(t)=k\}}+=\sum_{j=0}^{6}\sum_{i=0}^{23} \gamma_{j,k}\mathbbm{1}_{\{d(t)=k,h(t)=k\}}+X_i(t)
\end{align}
where $\mathbbm{1}$ is the indicator function. In (\ref{eq:mixedeffect}), coefficients $\alpha_j$ and $\beta_k$ capture the individual day-of-week and time-of-day effects, and $\gamma_{j,k}$ captures their joint effect. The stochastic process $X_i(t)$ captures the variation of arrival rate $M_i(t)$ from the value that is expected due to the fixed effects. We determine coefficients  $\alpha_j,\beta_k,$ and $\gamma_{j,k}$ using  least squares estimation.\footnote{We choose a least squares estimation  as it does not require any assumption about $X_i(t)$. If $X_i(t)$ are i.i.d, then the least square estimation is the same as maximum likelihood estimation. However, Figure \ref{fig:CF-R-27} suggests that there exists a dependency between $X_i(t)$ over time. } %We use a multiplicative model to represent arrival process $M_i(t)$. Specifically, in (\ref{eq:mixedeffect}) we use $\log M_i(t)$, instead of $M_i(t)$, since our analysis suggests that a \textit{multiplicative} model provides a better framework to capture the variation of arrival processes over time. 

Consider Figure \ref{fig:FT-27}, which provides a comparison between the Fourier Transformation (FT) of the aggregate arrival process $M(t)$ for parking location $27$ and the aggregate random effect $\sum_{i=1}^4X_i(t)$ after we subtract the estimated fixed effects. Consistent with the findings in Section \ref{sec:analysis-arrival}, the FT of $M(t)$ has distinct spikes over the frequencies that correspond to one-week and one-day periods, or higher multiples of them. By comparison, the FT of random effect $X(t)$ has much shorter spikes over these frequencies. Comparing the relative magnitude/power of the fixed and random effects, the values of $\sum_{i=1}^4|X_i(t)|^2$ is approximately $10\%$ of $\sum_{i=1}|M_i(t)|^2$. %The results suggest that a multiplicative model can better capture the periodic patterns due to  the day-of-week and time-of-day effects. Using the multiplicative model, the random effects $X_i(t)$ appear as a \textit{proportional} increase/decrease in arriving rate $M(t)$ that is correlated over time.        

Figure \ref{fig:CF-R-27} shows the ACF and PACF for the random effect $X_4(t)$ for population $4$ (long stops) for parking location $27$; we note that population $4$ has the highest significance in the prediction of the parking occupancy as they capture vehicles with long stopping times. A similar set of results for ACF and PACF is observed for other populations as well. Comparing with Figure \ref{fig:CF-27}, the results suggests that the estimated fixed effects due to time-of-day and day-of-week capture most of the cross-correlation between arrivals over time; this is consistent with the observation made above based on FT. 

Figure \ref{fig:CF-R-27} shows that the remaining random effects $X_i(t)$, $i=1,\cdots,4$  has a small value of autocorrelation over time, specifically, for small lags and lags around multiples of $24$. To capture such a structure in random effects process, we assume that $X_i(t)$, $i=1,\cdots,4$ follows a seasonal autoregressive integrated moving average model (SARIMA) \cite{shumway2017time}. Let $B$ denote the lag-1 operator and $\Delta_s$ denote the difference operator of period $s$; that it, $B\;X(t)=X(t-1)$ and $\Delta_s X(t)=X(t)-X(t-s)$. A SARIMA model with parameters $(p,d,q)\hspace*{-2pt}\times\hspace*{-2pt}(P,D,Q)_s$ is defined as
\begin{align}
\Phi(B^s)\phi(B)\Delta_s^D\Delta_1^d X(t)=\Theta(B^s)\theta(B)\epsilon(t),
\end{align}
where
\begin{align*}
&\phi(B)=1-\phi_1B-\phi_2B^2-\cdots-\phi_pB^p,\\
&\theta(B)=1-\theta_1B-\theta_2B^2-\cdots-\theta_qB^q,\\
&\Phi(B^s)=1-\Phi_1B^s-\Phi_2B^{2s}-\cdots-\Phi_PB^{Ps},\\
&\Theta(B^s)=1-\Theta_1B^s-\Theta_2B^{2s}-\cdots-\Theta_QB^{Qs},\\
\end{align*}
and $\epsilon(t)$ is a sequence of i.i.d normally distributed random variables; the value of coefficients $\phi_i,\theta_i,\Phi_i,\Theta_i$ are estimated using the training data. The SARIMA model described above presents an extension of the classical ARIMA$(p,d,q)$ model to include seasonal effects with period $s$ captured by AR and MA characteristic polynomials $\Phi(B^s)$ and $\Theta(B^s)$, and $D$ order of seasonal differentiation.

For model identification, we first use the Augmented Dickey-Fuller (ADF) test \cite{shumway2017time} and Kwiatkowski-Phillips-Schmidt-Shin (KPSS) test \cite{kwiatkowski1992testing} to determine the number of unit roots (i.e. integration order). Both tests result in the p-values of smaller than $0.01$ for all populations across all locations. This is consistent with the observation made in Figure \ref{fig:CF-R-27} where the autocorrelation is vanishing over time excluding the seasonal correlation present at multiples of $24$.  Therefore, we choose $d=0$. Moreover, we set $p=1$ and $q=0$ since the observed PAC for lag-1 is significantly higher than for lags larger than one; see Figure \ref{fig:CF-R-27}. To determine the seasonal parameters $(P,Q,D)_s$, we note that ACF in Figure \ref{fig:CF-R-27} shows a small but non-vanishing positive correlation for lags at multiples of $24$. Therefore, we set $s=24$, and choose an order $D=1$ for the seasonal difference along with a seasonal MA model of order $Q=1$ and set $P=0$. Our simulation for all population across all locations results in similar parameter selections for a SARIMA model.      
Consequently, we assume that every random effect $X_i(t)$ ,$i=1,..,4$ follows a SARIMA $(1,0,0)\times(0,1,1)_{24}$ model.

Figure \ref{fig:CF-RR-27} shows the ACF and PACF for residual error $\epsilon_4(t)$ for population $4$ with long stops for parking location $27$; similar results are obtained for other populations and parking locations. Given the negligible values of ACF and PACF, we adopt the mixed-effect model proposed above for the prediction of arrival rates $M_i(t)$, $=i,
\cdots,4$.

\textbf{Alternative models:} We note that a SARIMA model with additional AR and MA terms does not result in a significant performance improvement for the prediction of arrival rate $M_i(t)$ based on Akaike information criterion (AIC) and Bayesian information criterion (BIC) \cite{shumway2017time}. Therefore, to avoid over-fitting, we do not consider a more complex SARIMA model. We can also consider a vectorized SARIMA model to jointly predict all arrival rates $M_i(t)$, $i=1,\dots,4$, by utilizing the correlation between them. Table \ref{table:corr} shows the locations-average of Pearson correlation coefficient between the arrival rates $M_i(t)$, $i=1,\cdots,4$. We note that for $M_3(t)$ (normal stops) and $M_4(t)$ (long stops), which have the most impact on occupancy prediction, the average cross-correlation coefficients are relatively small. Moreover, our experimentation with a vectorized SARIMA does not result in a consistent improvement based on AIC and BIC across all locations. Similarly, our investigation suggests that a joint prediction of arrival rates for all parking locations does not lead to a performance improvement based on AIC and BIC.          

\vspace*{-5pt}
\parbox [t!]{0.48\textwidth }{
	\hspace*{-20pt}\includegraphics[width =1.15\linewidth ]{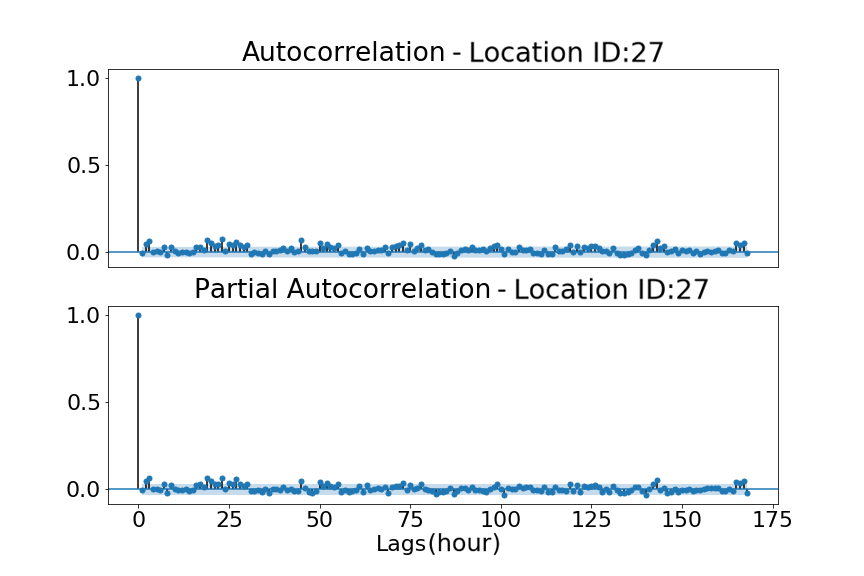}
	\captionof{figure}{The ACF (top plot) and PACF (bottom plot) for residual error $\epsilon_4(t)$ for population $4$ with long stops - location $27$.}
	\label{fig:CF-RR-27}
}
\hfill
\parbox [t!]{0.48\textwidth }{
	{\fontsize{9}{10}\hspace*{-20pt}\selectfont\begin{tabular}{|c|c|c|c|c|}
			\hline
			\hspace*{-7pt} Population \hspace*{-7pt}& \hspace*{-9pt}$\begin{array}{c}\text{Very Short} \\ (0\hspace*{-1pt}-\hspace*{-1pt}5) \hspace*{-1pt}\text{ min} \end{array}$ \hspace*{-10pt}& \hspace*{-10pt}$\begin{array}{c} \text{Short} \\ (5\hspace*{-1pt}-\hspace*{-1pt}25) \hspace*{-1pt}\text{ min} \end{array}$ \hspace*{-10pt}& \hspace*{-10pt} $\begin{array}{c} \text{Normal} \\ (25\hspace*{-1pt}-\hspace*{-1pt}360) \hspace*{-1pt}\text{ min} \end{array}$\hspace*{-10pt} & \hspace*{-10pt} $\begin{array}{c} \text{Long} \\ $>6$ \text{ hr} \end{array}$ \hspace*{-9pt}\\
			\hline
			\hspace*{-9pt}$\begin{array}{c}\text{Very Short} \\ (0\hspace*{-1pt}-\hspace*{-1pt}5) \hspace*{-1pt}\text{ min} \end{array}$ \hspace*{-11pt}& 1 & 0.27 & 0.09 & 0.14 \hspace*{-9pt}\\
			\hline
			\hspace*{-9pt}$\begin{array}{c}\text{Short} \\ (5\hspace*{-1pt}-\hspace*{-1pt}25) \hspace*{-1pt}\text{ min} \end{array}$ \hspace*{-11pt}& 0.27 & 1 & -0.02 & 0.03 \hspace*{-9pt}\\
			\hline
			\hspace*{-9pt}$\begin{array}{c}\text{Normal} \\ (25\hspace*{-1pt}-\hspace*{-1pt}360) \hspace*{-1pt}\text{ min} \end{array}$ \hspace*{-11pt}& 0.09 & -0.02 & 1 & 0.005 \hspace*{-9pt}\\
			\hline
			\hspace*{-9pt}$\begin{array}{c}\text{Long} \\ (>6) \text{ hr} \end{array}$ \hspace*{-11pt}& 0.14 & 0.03 & 0.005 & 1 \hspace*{-9pt}\\
			\hline
		\end{tabular}}
		\captionof{table}{The Pearson correlation coefficient (average over all locations) for arrival rates $M_i(t)$ and $M_j(t)$, $i,j=1,\cdots,4$.}
		\label{table:corr}
	}
		
An alternative approach for the prediction of arrival rates $M_i(t)$ $i=1,\dots,4,$ is to use a more sophisticated machine learning algorithm such as recurrent neural network (RNN). Specifically, a class of RNN called Long short-term memory (LSTM) networks that are proposed specifically for prediction in time-series can be utilized. Our limited experimentation with this class of neural networks did not result in significant improvement over the prediction accuracy obtained from the SARIMA model. We note that a more extensive study focusing on developing prediction algorithms that improve upon the proposed SARIMA model is an interesting direction for future research. However, as we discuss in Section \ref{sec:prediction}, the development of such prediction algorithms for arrival rates has a minimal effect on increasing the accuracy of prediction for parking occupancy. As we point out, a more effective approach is to develop a more accurate parametrization of time-varying service $G(t)$.       
  
% \red{(red add plots for arrival prediction performance?)}

%\input{Prediction}
\section{Occupancy Prediction}\label{sec:prediction}

Based on the framework proposed proposed in Sections \ref{sec:model-description}-\ref{sec:analysis-arima}, we design a probabilistic forecast of parking occupancy over time. We present two main types of predictions: 1) a microscopic prediction that keeps track of each parking spot separately and estimates the time-varying service time distribution using the decomposition of arrivals into four populations described above, and 2) a macroscopic prediction that is only based on the aggregate occupancy level and does not consider each parking spot separately. We discuss the advantage of each method and how they can be potentially improved utilizing more advanced machine learning techniques.

\subsection{Microscopic Prediction}

Given the decomposition of arriving vehicles into four populations, we can rewrite (\ref{eq:occupied}) as follows

\begin{align}
\mu_N(t)=\sum_{i=1}^4\mathbb{E}\{N_i\}(t)=\sum_{i=1}^4\int_{0}^{\infty} M_i(t-s)(1-G_i(s))ds. \label{eq:occupied-pop}
\end{align}

where $N_i(t)$ is a Poisson random variable denoting the number of spots occupied at time $t$ by vehicles from population $i$.

Equation (\ref{eq:occupied-pop}) provide an \textit{off-line} probabilistic forecast of occupancy as it does not use any real-time observation about vehicles already parked at time $t$. Using the mixed effect model described in Section \ref{sec:analysis-arima}, we can determine an \textit{on-line} probabilistic forecast of occupancy for a future time $t+\Delta t$ based on the \textit{current state} at $t$ as follows. Define the current state $Q(t)$ of a parking lot as the arrival times of all vehicles that are parked in the parking lot at time $t$ as 
\begin{align}
Q(t):=(\tau_{A(t)},\cdots,\tau_{A(t)-N(t)+1})
\end{align} where $\tau_{A(t)}$ and $\tau_{A(t)-N(t)+1}$ denote the arrival times  for the newest and oldest vehicle still parked in the parking lot, respectively.

A vehicle that is parked at $t+\Delta t$  either (i) arrived before time $t$, or (ii) arrived after time $t$. 
Define $N_o(t+\delta;t)$ as the number of vehicles parked at $t+\Delta t$ that arrive before time $t$. Similarly, let $N_n(t+\delta;t)$ denote the number of vehicles parked at $t+\Delta t$ that arrive after time $t$. We determine the probability distribution of $N_o(t+\delta;t)$ and $N_n(t+\delta;t)$ below.  

\textbf{(i)} Consider vehicle $i$ that arrives before time $t$, \textit{i.e.} $i\in\{A(t)-N(t)+1,\cdots,A(t)\}$. Then,
\begin{align}\label{eq:before-single}
\hspace*{-5pt}\mathbb{P}\{\text{veh. \hspace*{-1pt}$i$ \hspace*{-1pt}parked \hspace*{-1pt}at \hspace*{-1pt}$t\hspace*{-2pt}+\hspace*{-2pt}\Delta t$}\}
\hspace*{-2pt}&=\hspace*{-2pt}
\sum_{j=1}^4 \mathbb{P}\{\text{veh.\hspace*{-1pt} $i$ \hspace*{-1pt}parked\hspace*{-1pt} at\hspace*{-1pt} $t\hspace*{-2pt}+\hspace*{-2pt}\Delta t$}|\text{veh.\hspace*{-1pt} $i$ \hspace*{-1pt}$\in$\hspace*{-1pt} population\hspace*{-1pt} $j$}\}\mathbb{P}\{\text{veh.\hspace*{-1pt} $i$\hspace*{-1pt} $\in$\hspace*{-1pt} population\hspace*{-1pt} $j$}\}\nonumber\\
&=\sum_{i=1}^4 \frac{G_j(t+\Delta t-\tau_i)}{G_j(t-\tau_i)}\frac{G_j(t-\tau_i)\lambda_j(\tau_i)}{\sum_{k=1}^4 G_k(t-\tau_i)\lambda_k(\tau_i)}=\frac{\sum_{j=1}^4 G_j(t+\Delta t-\tau_i)\lambda_j(\tau_i)}{\sum_{k=1}^4 G_k(t-\tau_i)\lambda_k(\tau_i)}.
\end{align}

%Let $N_o(t+\delta;t)$ denote the number of vehicles parked at $t+\Delta t$ that arrive before time $t$. 
The probability distribution of $N_o(t+\delta;t)$ conditioned on current state $Q(t)$ is equal to the summation of $N(t)$ binomial random variables with parameters given by (\ref{eq:before-single}). Therefore,
\begin{align}
&\mathbb{E}\{N_o(t+\Delta t;t)|Q(t)\}=\sum_{i=A(t)-N(t)+1}^{A(t)} \frac{\sum_{j=1}^4 G_j(t+\Delta t-\tau_i)\lambda_j(\tau_i)}{\sum_{k=1}^4 G_k(t-\tau_i)\lambda_k(\tau_i)}\label{eq:N_o}
%&Var\{N_o(t+\Delta t;t)|Q(t)\}=\sum_{i=A(t)-N(t)+1}^{A(t)} \left(1-\frac{\sum_{j=1}^4 G_j(t+\Delta t-\tau_i)}{\sum_{k=1}^4 G_k(t-\tau_i)}\right)\frac{\sum_{j=1}^4 G_j(t+\Delta t-\tau_i)}{\sum_{k=1}^4 G_k(t-\tau_i)}
\end{align}

\textbf{(ii)} From  (\ref{eq:occupied-pop}), the number of vehicles parked at $t+\Delta t$ that arrive after $t$ is a Poisson random variable with mean
\begin{align}
\mathbb{E}\{N_n(t+\Delta t;t)|Q(t)\}=\mathbb{E}\{N_n(t+\Delta t;t)\}=\sum_{j=1}^4\int_0^{\Delta t} M_j(t+\Delta t-s)(1-G_j(s))ds.\label{eq:N_n}
\end{align} 
We note that the random variable $N_n(t+\delta;t)$ is independent of $Q(t)$.% Moreover, 
%\begin{align}
%Var\{N_n(t+\delta;t)\}=\mathbb{E}\{N_n(t+\Delta t;t)\}=\sum_{j=1}^4\int_0^{\Delta t} M_j(t+\Delta t-s)(1-G_j(s))ds.
%\end{align}

Combining (\ref{eq:N_o}) and (\ref{eq:N_n}), the probability distribution of $N(t+\Delta t)$ is given by the summation of $N_o(t+\Delta t;t)$ independent Bernoulli random variables and a Poisson random variable, where
\begin{align}
\hspace*{-5pt}\mathbb{E}\{N(t+\Delta t)|Q(t)\}=\sum_{j=1}^4\int_0^{\Delta t} M_j(t+\Delta t-s)(1-G_j(s))ds+\hspace*{-7pt}\sum_{i=A(t)-N(t)+1}^{A(t)}\hspace*{-7pt} \frac{\sum_{j=1}^4\hspace*{-1pt} G_j(t+\Delta t-\tau_i)\lambda_j(\tau_i)}{\sum_{k=1}^4 G_k(t-\tau_i)\lambda_k(\tau_i)}.\label{eq:N-mean}
\end{align}  

Using the law of large numbers, we can approximate the probability distribution of $N(t)$ for large enough $N_o(t+\delta;t)$ and $N_n(t+\delta;t)$ by a Gaussian distribution as
\begin{align}
\frac{N(t+\Delta t)-\mathbb{E}\{N(t+\Delta t)|Q(t)\}}{\sqrt{Var\{N(t+\Delta t)|Q(t)\}}}\approx \text{N}(0,1).\label{eq:micro-prob}
\end{align}

Estimating $Var\{N(t+\Delta t)|Q(t)\}$ is a more demanding task as it depends on (i) the intrinsic randomness present in the queuing model, and (ii) the estimation/prediction\footnote{For making a prediction $N(t+\Delta t;t)$ at time $t$ for time $t+\Delta t$, we need to estimate $M(\tau)$ and $G(\tau)$ for $\tau\leq t$ to determine $N_o(t+\Delta t;t)$, and predict $M(\tau)$ and $G(\tau)$ for $t\leq \tau\leq t+\Delta t$ to determine $N_n(t+\Delta t;t)$; see (\ref{eq:N-mean}). We note that for $\tau\leq t$ the observed number of arrivals is a realization of the a Poisson variable with rate $M(t)$ and is not necessary equal to $M(t)$.} error for these parameters based on the proposed parametrization of time-varying service time $G(t)$ and the SARIMA model for the arrival rates.
In the following, we first provide a model-based estimation of $Var\{N(t+\Delta t)|Q(t)\}$ based on point (i). Such an estimation of  $Var\{N(t+\Delta t)|Q(t)\}$ provides a lower bound on the accuracy of an arbitrary prediction algorithm. We then determine a numerical estimation of $Var\{N(t+\Delta t)|Q(t)\}$ that captures both points (i) and (ii) for the specific prediction algorithm we proposed here.

\textbf{Lower Bound:}
Even when all parameters of $M(t)/G(t)/\infty$ are perfectly known for all time, the queuing model has an intrinsic randomness in the realization of the number of arrivals and service times. That is, the number of arrivals at every time interval is a Poisson random variable with parameter $M(t)$, and service time for an vehicle arriving at time $t$ is a random variable with CDF $G(t)$. Ignoring the estimation/prediction error for these parameters, we can compute $Var\{N(t+\Delta t)|Q(t)\}$ as
\begin{align}
Var_{\text{LB}}\{N(t+\Delta t)|Q(t)\}	=&Var_{\text{LB}}\{N_o(t+\Delta t;t)|Q(t)\}+Var_{\text{LB}}\{N_n(t+\delta;t)\}\nonumber\\
=&\sum_{i=A(t)-N(t)+1}^{A(t)} \left(1-\frac{\sum_{j=1}^4 G_j(t+\Delta t-\tau_i)\lambda_j(\tau_i)}{\sum_{k=1}^4 G_k(t-\tau_i)\lambda_k(\tau_i)}\right)\frac{\sum_{j=1}^4 G_j(t+\Delta t-\tau_i)\lambda_j(\tau_i)}{\sum_{k=1}^4 G_k(t-\tau_i)\lambda_k(\tau_i)}\nonumber\\&+\sum_{j=1}^4\int_0^{\Delta t} M_j(t+\Delta t-s)(1-G_j(s))ds\label{eq:N-var}
\end{align}
where the first term denotes the variance of $N_o(t+\Delta t;t)$ ($N(t)$ binomial random variables) and the second term denotes the variance of $N_n(t+\Delta t;t)$  (a Poisson random variable). 
  
 %provides an expected upper bound on the prediction accuracy of any prediction method even if we have perfect knowledge/prediction of all model parameters. \blue{Why?} 
We note that in deriving (\ref{eq:N-var}), we only assumes that Assumptions \ref{assump-poisson}-\ref{assump-infinite} hold, which we empirically verified in Section \ref{sec:model-verification}. Therefore, we can use the value of (\ref{eq:N-var}) as a lower bound on prediction errors to evaluate the performance of different prediction algorithms. 

%
%\begin{align}
%Var\{N(t+\Delta t)|Q(t)\}	=&Var\{N_o(t+\Delta t;t)|Q(t)\}+Var\{N_n(t+\delta;t)\}\nonumber\\
%=&\sum_{i=A(t)-N(t)+1}^{A(t)} \left(1-\frac{\sum_{j=1}^4 G_j(t+\Delta t-\tau_i)\lambda_j(\tau_i)}{\sum_{k=1}^4 G_k(t-\tau_i)\lambda_k(\tau_i)}\right)\frac{\sum_{j=1}^4 G_j(t+\Delta t-\tau_i)\lambda_j(\tau_i)}{\sum_{k=1}^4 G_k(t-\tau_i)\lambda_k(\tau_i)}\nonumber\\&+\sum_{j=1}^4\int_0^{\Delta t} M_j(t+\Delta t-s)(1-G_j(s))ds\label{eq:N-var}
%\end{align}

\begin{figure}[t!]
	\begin{minipage}{0.48\textwidth}
		\centering
		\hspace*{-20pt}\includegraphics[width=1.15\textwidth]{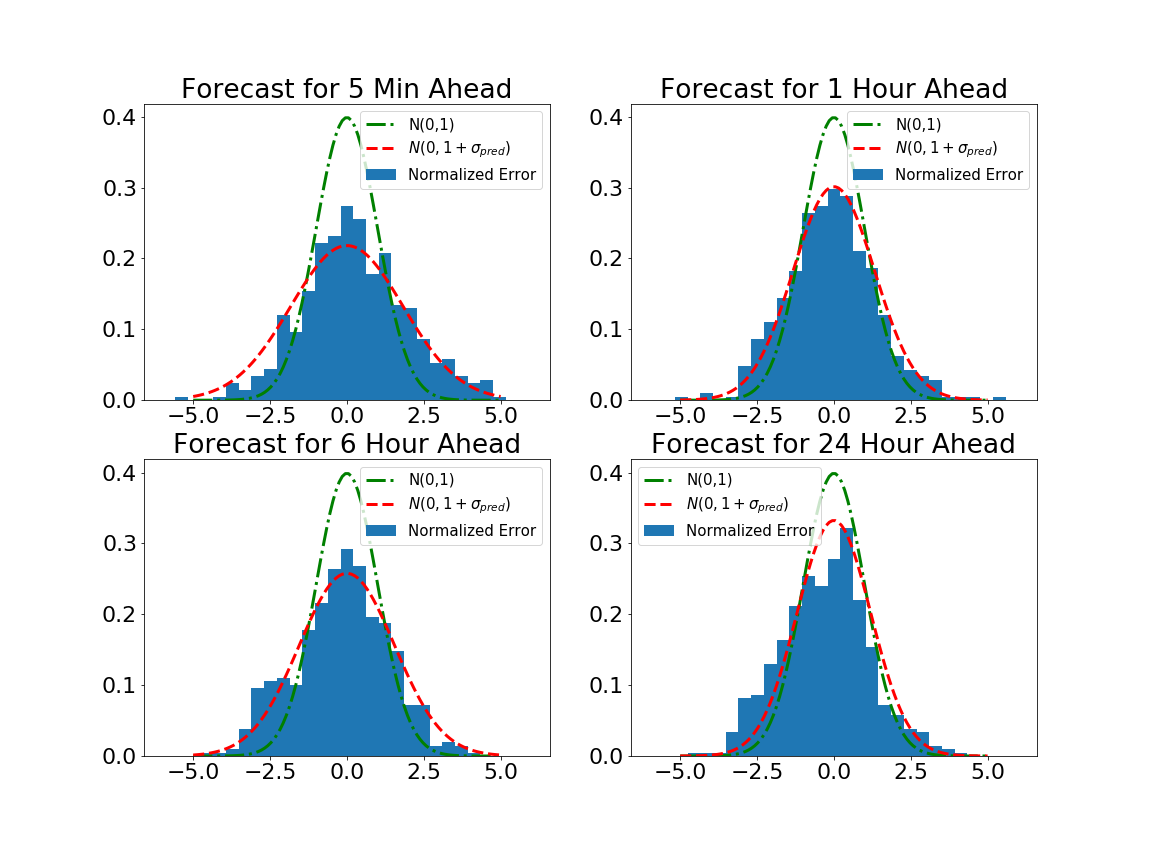}
		\caption{ Microscopic prediction: the probabilistic forecast of normalized parking occupancy $\frac{N(t+\Delta t)-\mathbb{E}\{N(t+\Delta t)|Q(t)\}}{\sqrt{Var_{\text{LB}}\{N(t+\Delta t)|Q(t)\}}}$ using microscopic prediction method (in blue), the estimated normal Gaussian distribution $N(0,1+\sigma^2_{\text{pred}})$ (in red), and a normal Gaussian distribution $N(0,1)$ (in green) - parking location $24$.  }
		\label{fig:online-31}
	\end{minipage}%
	\hspace*{10pt}
	\begin{minipage}{0.48\textwidth}
		\centering
		\hspace*{-20pt}\includegraphics[width=1.15\textwidth]{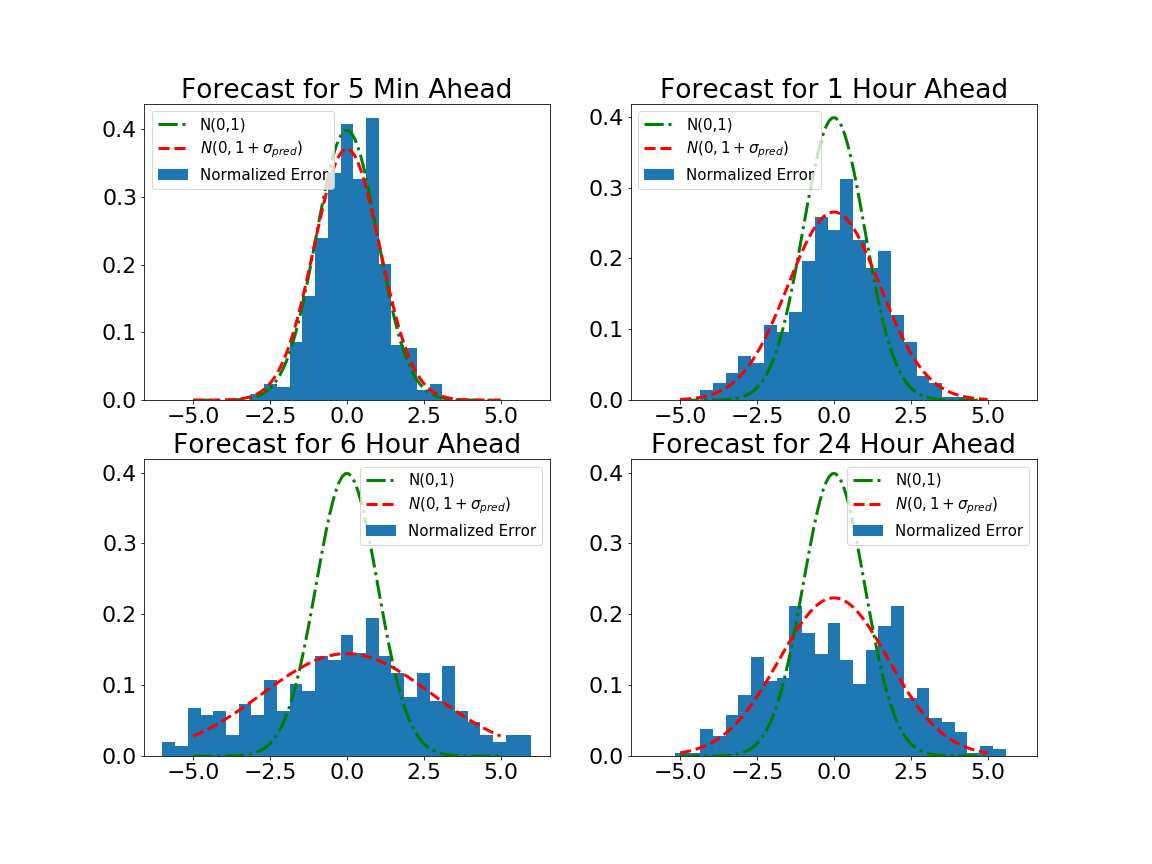}
		\caption{ Microscopic prediction: the probabilistic forecast of normalized parking occupancy $\frac{N(t+\Delta t)-\mathbb{E}\{N(t+\Delta t)|Q(t)\}}{\sqrt{Var_{\text{LB}}\{N(t+\Delta t)|Q(t)\}}}$ using microscopic prediction method (in blue), the estimated normal Gaussian distribution $N(0,1+\sigma^2_{\text{pred}})$ (in red), and a normal Gaussian distribution $N(0,1)$ (in green) - parking location $27$.  }
		\label{fig:online-27}
	\end{minipage}
\end{figure}%

\textbf{Variance Estimation:}
%Given the current state $Q(t)$, equation (\ref{eq:N-var}) determines the expected variance of $N(t+\Delta t)$ due to the intrinsic randomness present in the realizations of the number of arrivals and service times in the $M(t)/G(t)/\infty$ model provided that all model parameters $M_i(t)$, $i=1,\ldots,4$, and $G_i(t)$ are perfectly known for all times. 
To predict $N(t+\Delta t;t)$, we need to predict the model parameters $M_i(t)$, $i=1,\ldots,4$, and $G(t)$, which comes with its own prediction error adding to the intrinsic error given by (\ref{eq:N-var}). Let $\sigma_{\text{pred}}^2:=\frac{Var\{N(t+\Delta t)|Q(t)\}}{Var_{\text{LB}}\{N(t+\Delta t)|Q(t)\}}-1$ denote the additional (relative) prediction errors that arises due to the parameters estimation/prediction. 
Since we do not have a \textit{``true model''} that captures the time variation of $M(t)$ and $G(t)$, we cannot provide a model-based approach to determine $\sigma^2_{\text{pred}}$. Nevertheless, we can estimate $\sigma^2_{pred}$ (or $Var\{N(t+\Delta t)|Q(t)\}$) numerically based on the historical data. We demonstrate such a numerical estimation of $\sigma^2_{\text{pred}}$ below.
%provides an expected upper bound on the prediction accuracy of any prediction method even if we have perfect knowledge/prediction of all model parameters. \blue{Why?} 
%Therefore, we can use the value of (\ref{eq:N-var}) as a lower bound on prediction errors to evaluate the performance of different prediction algorithms.
 
Equations (\ref{eq:N-mean}) and (\ref{eq:micro-prob}) provide a probabilistic forecast of parking occupancy for $t+\Delta t$ based on the current state at $t$; we refer to this method as \textit{the microscopic prediction} as $Q(t)$ keeps track of each parking spot. We can also form a point prediction of parking occupancy using the microscopic method, which is both the maximum likelihood estimator and the least squares error estimator, namely $\mathbb{E}\{N(t)|Q(t)\}$. 

Figures \ref{fig:online-31} and \ref{fig:online-27} show the empirical distribution of normalized prediction error $\frac{N(t+\Delta t)-\mathbb{E}\{N(t)|Q(t)\}}{\sqrt{Var_{\text{LB}}\{N(t+\Delta t)|Q(t)\}}}$ (in blue) compared to the estimated normal Gaussian distribution $N(0,1+\sigma^2_{\text{pred}})$ (in red), and a normal Gaussian distribution $N(0,1)$ (in green), for parking locations $24$ and $27$. We estimate the parameters of our model based on the data collected between February, 02, 2017 to April, 08, 2018. The empirical distribution is generated by aggregating the occupancy prediction for the next $\Delta t\in\{5\text{ min},1\text{ hr},6\text{ hr},24\text{ hr}\}$ given the current state $Q(t)$ at the beginning of every hour during the three weeks time interval between April 16, 2018, to May 06, 2018, resulting in $504$ prediction samples for every horizon. We use $168$ prediction samples  determined at the beginning of every hour during the one week time interval between April 09, 2016 to April 16, 2018 to estimate $\sigma^2_{\text{pred}}$ for each time horizon $\Delta t\in\{5\text{ min},1\text{ hr},6\text{ hr},24\text{ hr}\}$.

\begin{figure}[t!]
	\begin{minipage}{0.48\textwidth}
		\centering
		\hspace*{-20pt}\includegraphics[width=1.15\textwidth]{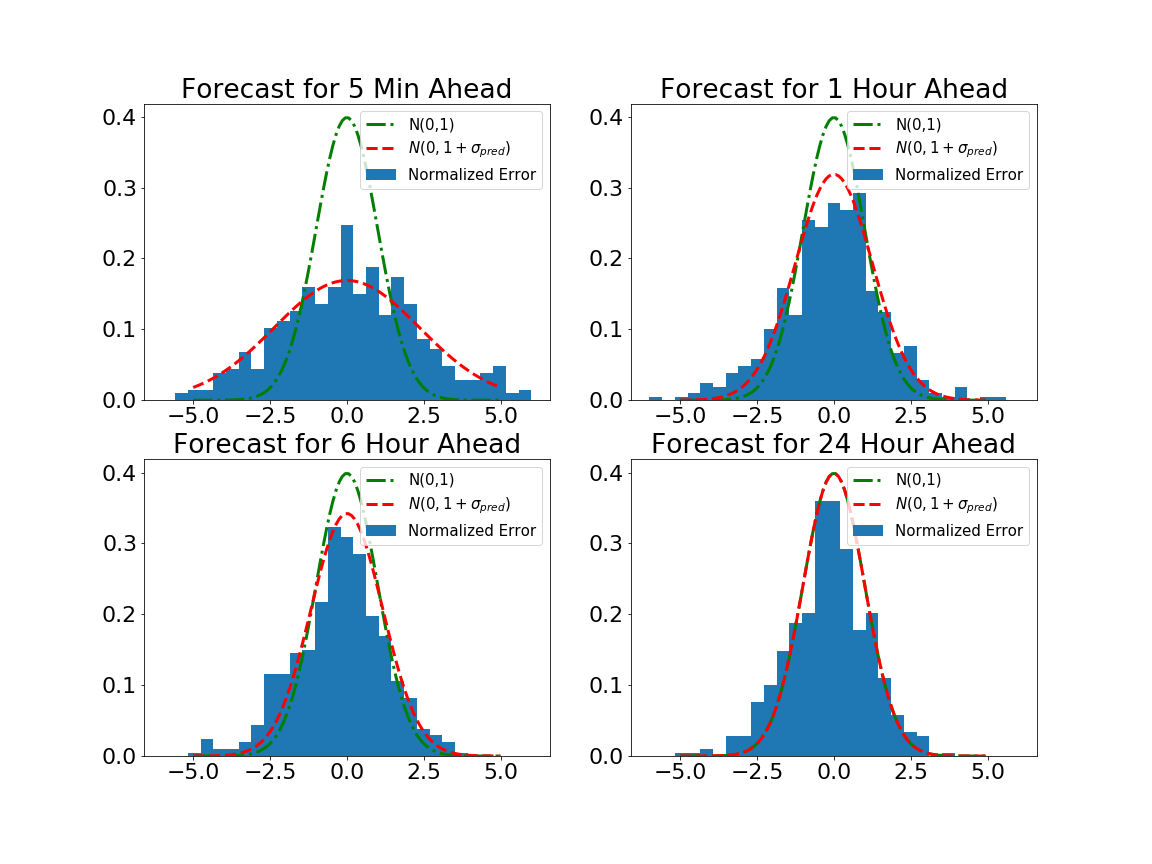}
		\caption{ Macroscopic prediction: the probabilistic forecast of normalized parking occupancy $\frac{N(t+\Delta t)-\mathbb{E}\{N(t+\Delta t)|Q(t)\}}{\sqrt{Var_{\text{LB}}\{N(t+\Delta t)|Q(t)\}}}$ using the macroscopic prediction method (in blue), the estimated normal Gaussian distribution $N(0,1+\sigma^2_{\text{pred}})$ (in red), and a normal Gaussian distribution $N(0,1)$ (in green) - parking location $24$. }
		\label{fig:offline-31}
	\end{minipage}%
	\hspace*{10pt}
	\begin{minipage}{0.48\textwidth}
		\centering
		\hspace*{-20pt}\includegraphics[width=1.15\textwidth]{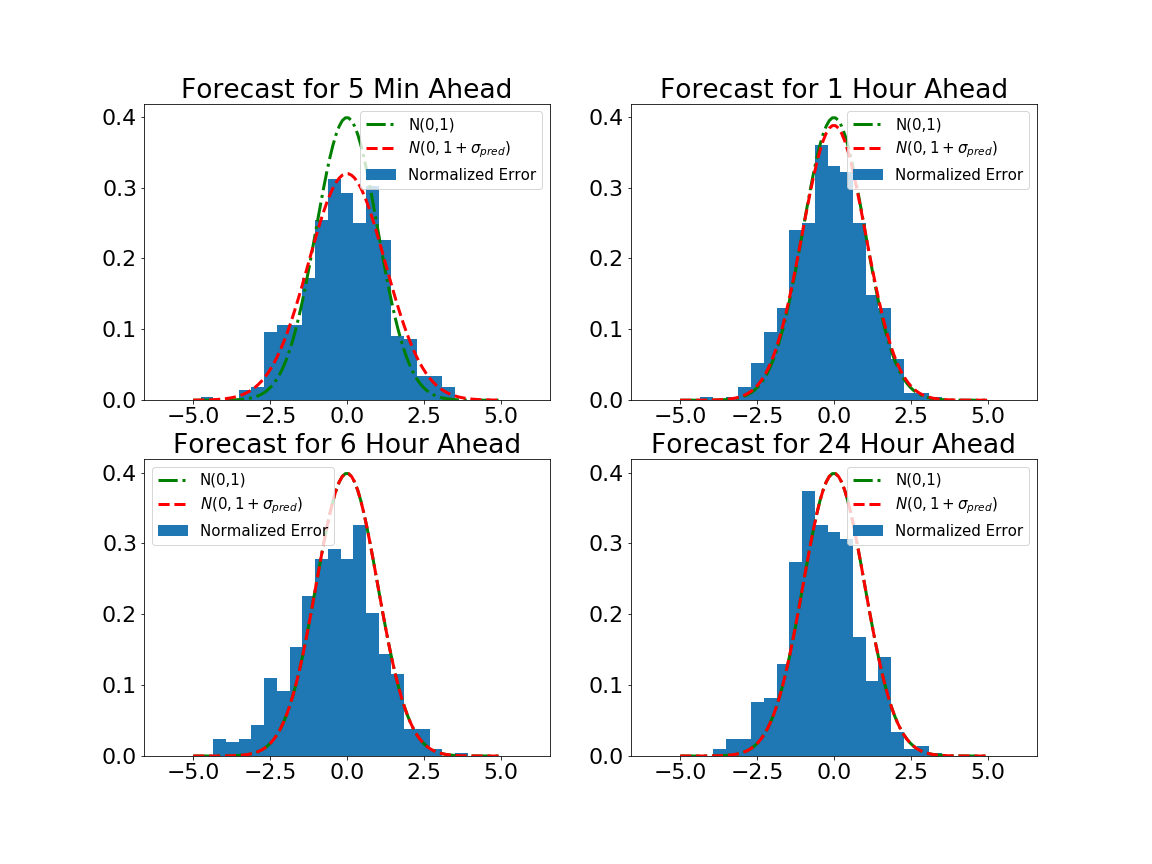}
		\caption{ Macroscopic prediction: the probabilistic forecast of normalized parking occupancy $\frac{N(t+\Delta t)-\mathbb{E}\{N(t+\Delta t)|Q(t)\}}{\sqrt{Var_{\text{LB}}\{N(t+\Delta t)|Q(t)\}}}$ using the macroscopic prediction method (in blue), the estimated normal Gaussian distribution $N(0,1+\sigma^2_{\text{pred}})$ (in red), and a normal Gaussian distribution $N(0,1)$ (in green) - parking location $27$. }
		\label{fig:offline-27}
	\end{minipage}
\end{figure}%

\subsection{Macroscopic Prediction}

The dependency of the microscopic prediction, given by (\ref{eq:N-mean}) and (\ref{eq:micro-prob}), on the current state $Q(t)$ vanishes as $\Delta t$ grows. That is, for large $\Delta t$, $N(t+\Delta)$ is approximately a Poisson random variable with mean $\int_0^{\Delta t} M(t+\Delta t-s)(1-G(s))ds$. Moreover, assuming $G(s)\rightarrow 0$ for $|s|\rightarrow \infty$, for large enough $\Delta t$, we can approximate the expected value of this Poisson random variable as
\begin{align*}
\int_0^{\Delta t} M(t+\Delta t-s)(1-G(s))ds\approx \int_0^{\infty} M(t+\Delta t-s)(1-G(s))ds:=N_{\text{macro}}(t+\Delta t).
\end{align*}

Therefore, to determine a probabilistic forecast for large  $\Delta t$, one only needs to predict the random process $N_{\text{macro}}(t+\Delta t)$; we call this probabilistic forecast \textit{the macroscopic prediction} as it only keeps track of aggregate occupancy level and is independent of current state $Q(t)$. 

The main advantage of the macroscopic prediction, compared to the microscopic prediction, is that $N_{\text{macro}}(t+\Delta t)$ can be predicted directly. This is different from the microscopic prediction where we need to first predict $M(t)$ and $G(t)$ separately by decomposing the arriving vehicles into four populations so as to approximate $G(t)$. Therefore, for large $\Delta t$, where the effect of current state $Q(t)$ is negligible, the macroscopic prediction offers a more accurate forecast by avoiding the prediction errors in forecasting $M(t)$ and $G(t)$.

For large parking lots, the Poisson random variable $N_{\text{macro}}(t+\Delta t)$ can be approximated by a Gaussian random variable. We can utilize various machine learning algorithms to predict  $\mathbb{E}\{N_{\text{macro}}(t+\Delta t)\}$ minimizing the sum of squared prediction errors. Moreover, we can numerically estimate $Var\{N(t+\Delta t)|Q(t)\}$ (or $\sigma_{\text{pred}}^2$) using historical data following an approach similar to the one for the microscopic prediction method. 

%We can utilize various machine learning algorithms to predict  $N_{\text{macro}}(t+\Delta t)$. %We note that for large parking lots, where the Poisson random variable $N_{\text{macro}}(t+\Delta t)$ can be approximated by a Gaussian random variable. We note that under the Gaussian approximation, to determine a point-wise prediction the outcome of a least squares estimator is approximately identical to that of a  maximum likelihood estimator. 
%We can use various prediction algorithms to generate a point-wise prediction of $N_{\text{macro}}(t+\Delta t)$. 

To provide a meaningful comparison of the two prediction methods, we use a prediction technique similar to the one used for the arrival rate prediction in Section \ref{sec:analysis-arima}. That is, we consider  a mixed effect model, with the time of the day and weekday effects, along with a simple SARIMA $(1,0,0)\times(0,1,1)_{24}$ model. We do not include a detail description of the steps that leads to choosing these parameters for the SARIMA model for $N_{\text{macro}}(t+\Delta t)$ as it follows a similar argument as the one presented in Section \ref{sec:analysis-arima}. %Moreover, we can numerically estimate $Var\{N(t+\Delta t)|Q(t)\}$ (ir $\sigma_{\text{pred}}^2$) using historical data following an approach similar to the one for microscopic prediction method. 

Figures \ref{fig:offline-31} and \ref{fig:offline-27} show the empirical distribution of $\frac{N(t+\Delta t)-N(t+\Delta t)}{\sqrt{Var_{\text{LB}}\{N(t+\Delta t)|Q(t)\}}}$ for parking locations $24$ and $27$, generated using the same dataset used for Figures \ref{fig:online-27} and \ref{fig:online-31}. 
Comparing Figures \ref{fig:online-31}-\ref{fig:offline-27}, the microscopic prediction offers a more accurate probabilistic forecast than the macroscopic prediction for small $\Delta t$, while the macroscopic method does better for large $\Delta t$. This is consistent with the argument given above for the advantage of the macroscopic prediction for large $\Delta t$. %However, we note that for small $\Delta t$, where the current state $Q(t)$ has a significant effect of occupancy level prediction the off-line prediction has a worse performance than that of on-line prediction.    

Figures \ref{fig:prediction-31} and \ref{fig:prediction-27} provide a comparison of the microscopic and macroscopic methods for parking locations $24$ and $27$, based on root mean square error (RMSE),  using the same time interval for training and evaluation as above. Consistent with the argument above, the microscopic method offers a better prediction accuracy for short forecast horizons, while the macroscopic method does better for longer forecast horizons.

\subsection{Comparison and Discussion}

The prediction performance of the microscopic method proposed above can be potentially improved by (i) utilizing a more complex algorithm, compared to a simple SARIMA, to predict the arrivals of the four populations, or alternatively (ii) to use a more sophisticated algorithm to predict the time-varying service time distribution function $G(t)$ rather than the parameterization of $G(t)$ proposed in Section \ref{sec:analysis} and writing it as weighted summation of four time-invariant functions $G_j,j=1,\cdots,4$. A comprehensive study of such improvements is an interesting research direction; however, we leave them for future work since the simple prediction techniques used above produce prediction errors that are reasonably close to the characterized expected lower bound.
Nevertheless, we briefly investigate the potential of such improvements, and compare the resulting  prediction performances. Moreover, we provide a comparison of our method with the queuing-based method proposed in \cite{xiao2018how}. %The results of this comparison appears in Figures \ref{fig:prediction-27} and \ref{fig:prediction-31}. 

\subsubsection{Improving the Microscopic Method:}  

Our analysis suggests that that developing a better prediction algorithm for service time distribution $G_t(s)$ has a higher potential impact compared to developing a better prediction algorithm for arrivals. To estimate an upper bound on the performance improvement achievable by utilizing a more sophisticated technique for arrival prediction, we consider a version of microscopic method where we assume that the arrival prediction is prefect.

\begin{figure}[t!]
	\begin{minipage}{0.48\textwidth}
		\centering
		\hspace*{-20pt}\includegraphics[width=1.15\textwidth]{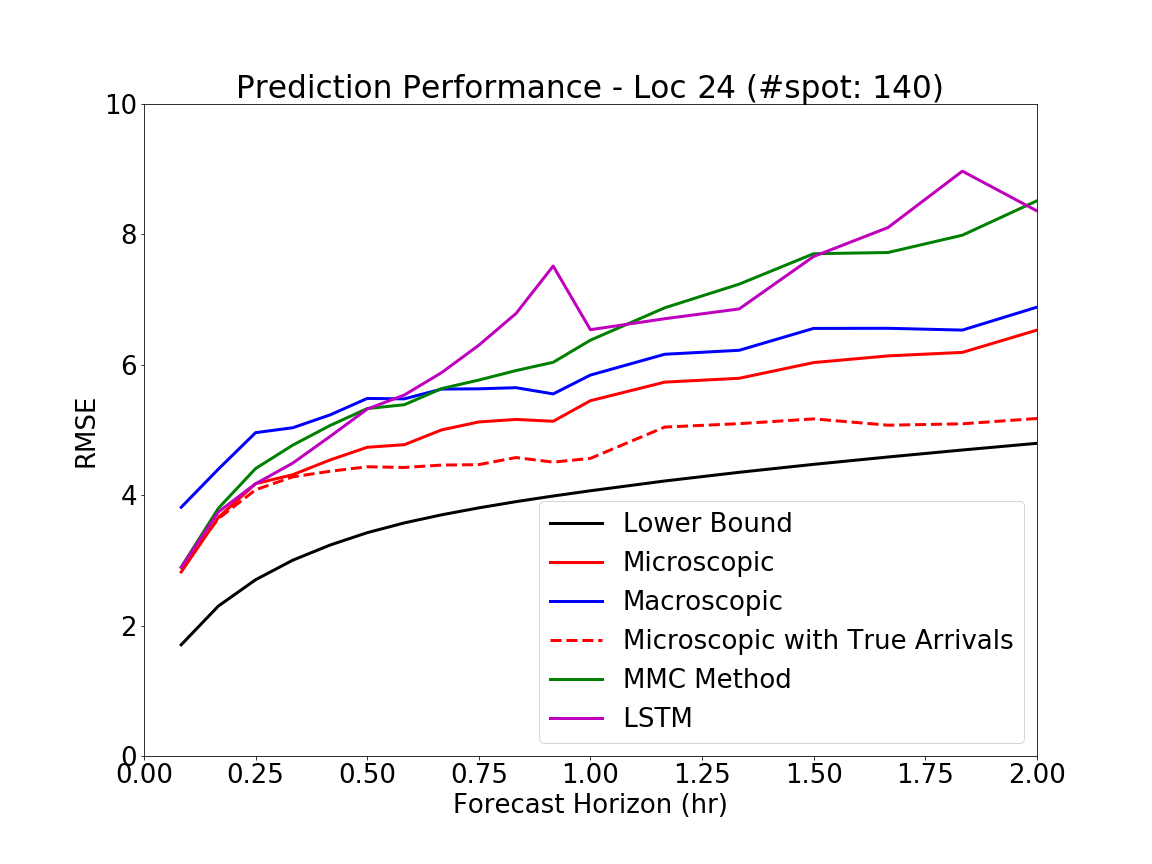}
		\hspace*{-20pt}\includegraphics[width=1.15\textwidth]{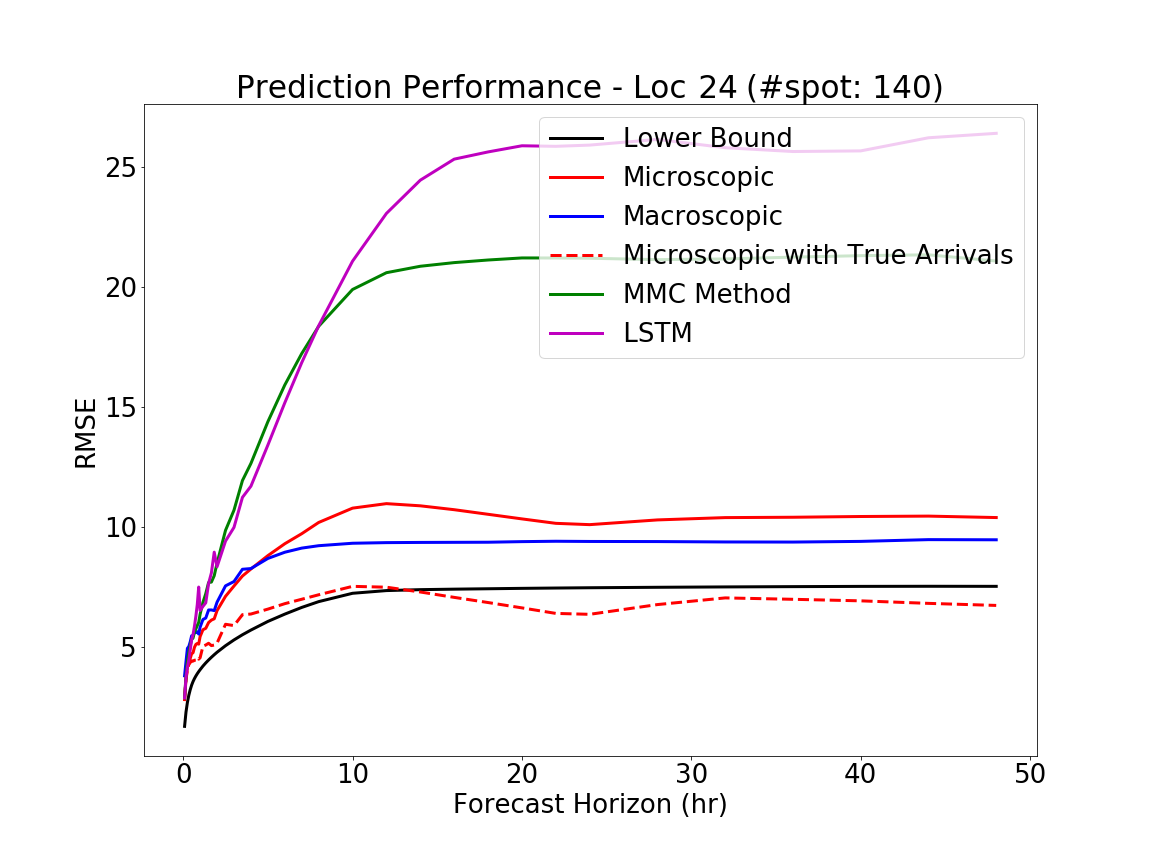}
		\caption{The accuracy of different prediction methods in terms of RMSE for forecast horizon $\Delta t\in[0,48]$ hours - parking location $24$}
		\label{fig:prediction-31}
	\end{minipage}%
	\hspace*{10pt}
	\begin{minipage}{0.48\textwidth}
		\centering
		\hspace*{-20pt}\includegraphics[width=1.15\textwidth]{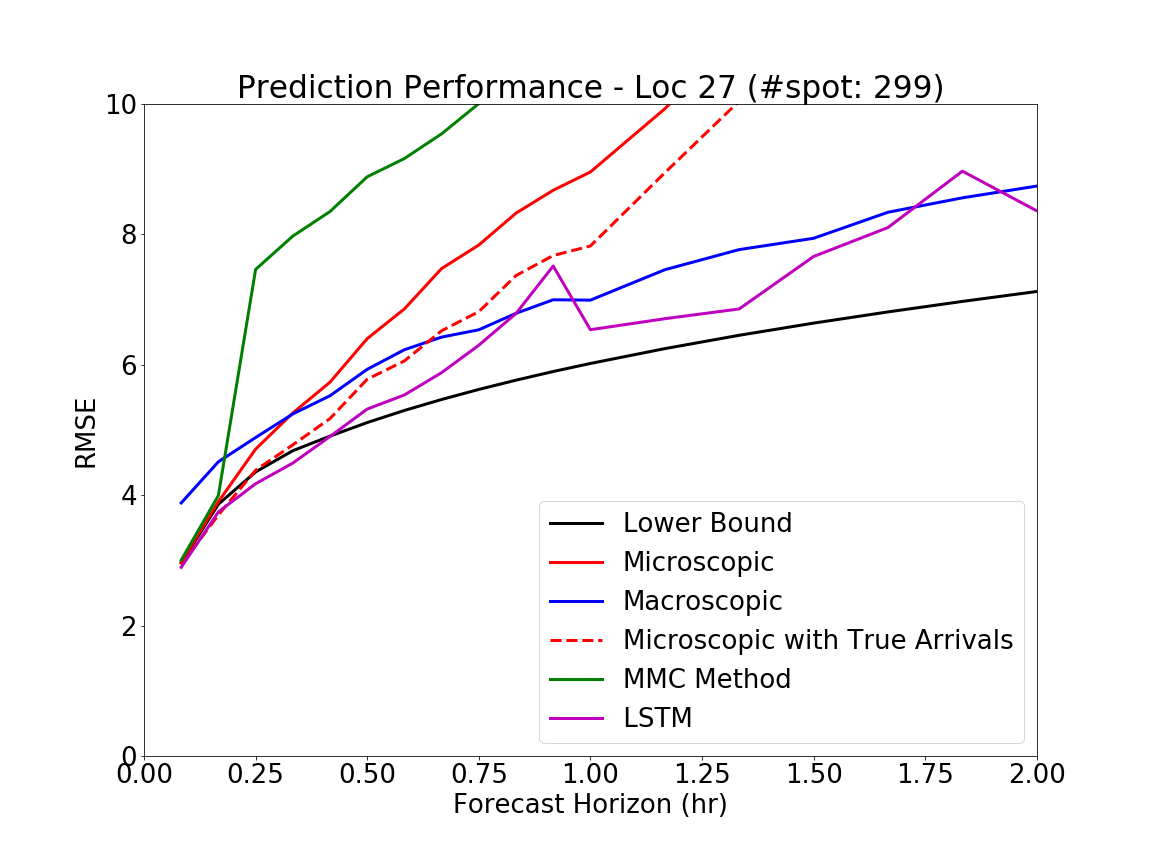}
		\hspace*{-20pt}\includegraphics[width=1.15\textwidth]{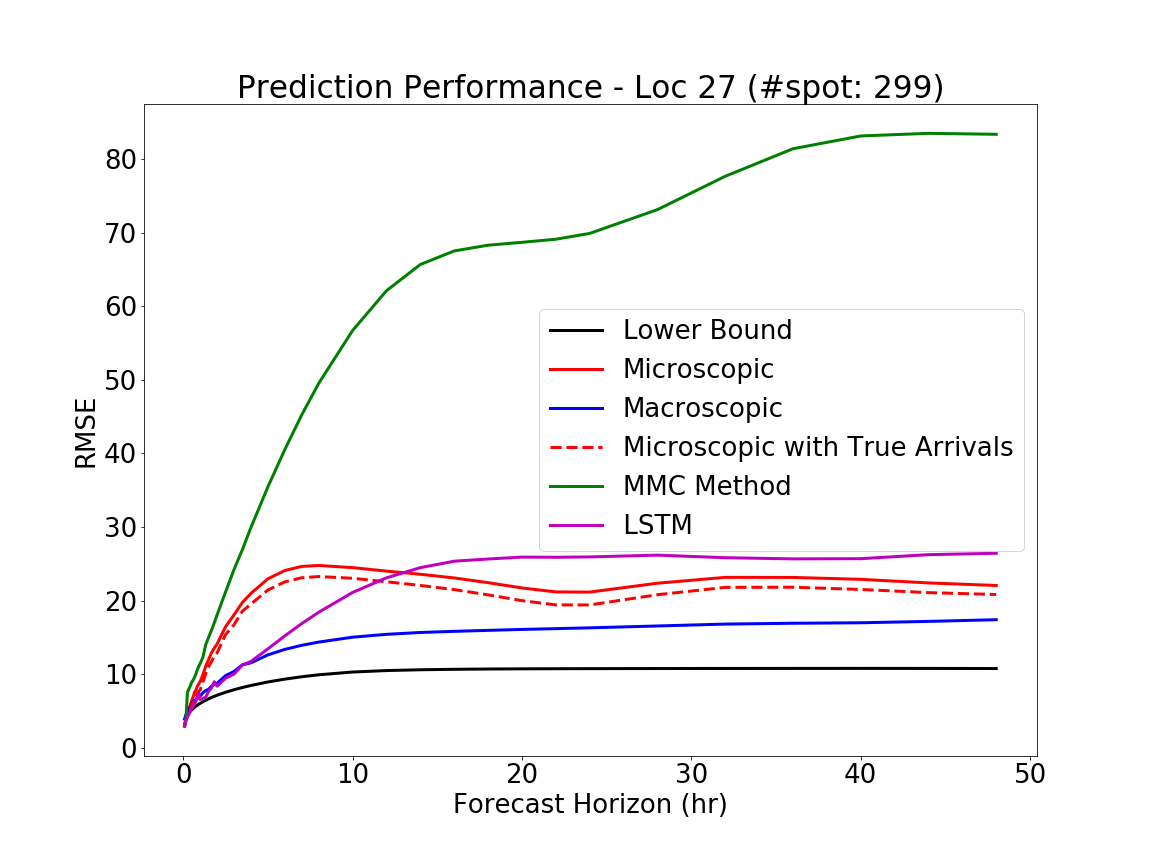}
		\caption{The accuracy of different prediction methods in terms of  RMSE for forecast horizon $\Delta t\in[0,48]$ hours - parking location $27$}
		\label{fig:prediction-27}
	\end{minipage}
\end{figure}

Figures \ref{fig:prediction-31} and \ref{fig:prediction-27} compare the root mean square error (RMSE) of the microscopic prediction method where we predict the arrival rates using the SARIMA model and the case where we have a perfect prediction of arrivals i.e. the arrival rates for all four population are known. The result suggests that any improvement in the prediction of arrival rates has limited effects on the accuracy of parking occupancy prediction, and the main factor contributing to the prediction error in the microscopic method is the error due to the prediction of $G(t)$. Therefore, an interesting future research direction is to develop an alternative algorithm for the prediction of distribution function $G(t)$.

\subsubsection{Improving the Macroscopic Method:}   

In the macroscopic method, we use a simple SARIMA model to predict the total occupancy level. There are many alternative machine learning algorithms one can choose instead of the SARIMA model. However, unlike the existing works in the literature that utilize model-free prediction algorithms to generate a point prediction, within the framework of the macroscopic method, we can utilize these algorithms to form a probabilistic forecast thanks to the modeling component of our methods. We do not aim to propose a prediction algorithm that is ``best'' for the macroscopic method in this paper; the choice of a ``best'' algorithm is potentially subject to various characteristics of the specific dataset we study. 

Nevertheless, we provide a comparison of the simple SARIMA model proposed above and an LSTM recurrent neural network  with $2$ hidden layers, each with 48 cells, an input window size of $24$, and $0.05$ dropout rate. We note that during the training for both the SARIMA and LSTM model, we minimize the prediction error for the next time ($5$ minutes ahead); the longer forecast horizons are generated by iteratively using the predicted output as the input for the next time. Moreover, we use the LSTM model to predict the random effect $X_i(t)$ after we extract the time-of-day and day-of-week effects (see (\ref{eq:mixedeffect})); our experimentation with the original data without subtracting these seasonal effects resulted in a significantly lower prediction performance. 

Figures \ref{fig:prediction-31} and \ref{fig:prediction-27} provide a comparison of these two methods for parking location $24$ and $27$. The LSTM results in a better prediction performance for short forecast horizons. This is consistent with the fact that the LSTM provides a richer function space compared to a simple SARIMA model. However, for larger forecast horizon the performance of LSTM deteriorates faster than that does SARIMA. This is because the explicit model structure of the SARIMA model provides an indirect learning regularization, which exhibits itself in a slower performance degradation for longer forecast horizon.  Our results suggest that implementing a more sophisticated prediction algorithm that provides a richer functional space, such as neural networks, must be done with caution considering an appropriate regularization method that ensures the generalizability of the results to longer forecast horizons.\footnote{An alternative approach is train a separate neural network for every forecast horizon. However, such an approach is computationally very expensive.}     

\subsubsection{Discussion and Comparison with $M/M/C$ Method:}

The lower bound estimated based on the $M_t/G_t/\infty$ model enables us to evaluate the relative performance of the microscopic and macroscopic method for every forecast horizon. The prediction errors of both methods remain approximately less than two times the lower bound. These performances appear to be satisfactory considering the fact that the lower bound does not include any error in the prediction of the arrivals rate $M(t+\Delta t)$ or service time distribution function $G(t+\Delta t)$.

As discussed in Section \ref{sec:relatedlit}, the work of \cite{xiao2018how} follows an approach that is closest to our work. We provide a comparison of our methods with the queuing based method proposed in \cite{xiao2018how}. The authors in \cite{xiao2018how} consider a $M/M/C$ queuing model to capture the parking dynamics. Similar, to our approach, they consider a case where the parking lot does not get completely full. Since the main objective in \cite{xiao2018how} is to form a short term prediction, they consider a queuing model with a constant arrival rate $\lambda$ and exponential distribution service time with parameter $\mu$. In a $M/M/C$ queue, the expected occupancy level $N(t+\Delta t)$ at $t+\Delta t$, given $N(t)$, is given by \begin{align}
N(t+\Delta t)=e^{-\lambda \Delta t}(N(t)-\frac{\lambda}{\mu})+\frac{\lambda}{\mu},\label{eq:MMC}
\end{align} 
where $\lambda$ and $\mu$ denote the fixed arrival and departure rates. We follow the approach in \cite{xiao2018how} and aggregate the data during one hour intervals with similar arrival and departure rates. According to the analysis in Section \ref{sec:analysis}, we consider the observed data for every pair (hour, weekday) to have similar parameters $(\lambda, \mu)$. We estimate $(\lambda, \mu)$ for every (hour, weekday) pair using the nonlinear regression model given by (\ref{eq:MMC}). We note that for forecast horizons that are longer than one hour we iteratively use equation (\ref{eq:MMC}), each time predicting  one hour ahead using the estimated $(\lambda,\mu)$ appropriate for that time. Therefore, we effectively extend the $M/M/C$ queuing model to a $M_t/M/C$ model with time-varying arrival rate for forecast horizon longer than one hour.    

The results show that the method proposed proposed \cite{xiao2018how} provides a satisfactory prediction performance for short forecast horizons, which is comparable to that of the microscopic method with SARIMA and the macroscopic method with LSTM; this is consistent with the results reported in \cite{xiao2018how}. However, as the forecast horizon increases the performance of it deteriorate significantly. This observation can be explained as follows. As we showed in Section \ref{sec:analysis}, the service time does not follow an exponential distribution. However, for short time horizons (<$1$ hour) the $M/M/C$ queuing based method provides the best functional estimate of the occupancy dynamics from the family of functions given by (\ref{eq:MMC}). Therefore, for short time horizon the prediction performance is satisfactory even though the parking dynamics is different from the $M/M/C$ model. However, as the forecast horizon increases, the prediction error due to the such model mismatch increases. 

%There are a few possible modifications to the on-line and off-line methods 
%We consider a few possible modifications to the on/off-line method proposed above, and briefly discuss their potential benefits. First, we can investigate an alternative algorithm for the prediction of arrivals for the on-line prediction methods. As we discussed, in Section \ref{sec:analysis-arima} our analysis of a more complex SARIMA model, or a LSTM neural network did not lead to a significant improvement in prediction accuracy. However, we argue that          

%\input{Conclusion}
\section{Concluding Remarks}\label{sec:conclusion}

We proposed a queuing model to capture parking dynamics. We provided a formal verification of all assumptions  underlying the queuing model using real data. Our work provides a formal ground for many analytic and policy design works that use a variation of the queuing model to study parking problems. Moreover, the queuing model and the verification method proposed in this paper present a framework that can be used and expanded for the study of parking problems in other settings. More specifically, for off-street parking in CBD, our framework can be applied without significant change if the parking lot is not saturated. For off-street parking lots that reach their full capacity regularly, one can verify the modeling assumptions only using the data during which the parking lot is not full. However, estimating the censored demand during the times when the parking lot is full is a challenging task and needs further investigation. 

A similar approach can be pursued for the study of on-street parking problems by treating each parking block as a separate parking lot. However, there are a few additional challenges in  on-street parking problems. First, most on-street parking areas in CBD are congested during work hours. This makes the estimation of arrival rates a very challenging task. The second challenge in highly congested on-street parking areas is the demand spillover to the neighboring areas during peak hours. This requires a joint model of arrival rates for geographically close parking blocks taking into account such spillovers. While our framework does not address these additional challenges, our results provide a first step toward the study of parking problems in these more complex environments.

In addition to the modeling part of this paper, we also proposed two prediction methods to provide a probabilistic forecast of parking occupancy level. We characterized an expected lower bound on prediction accuracy that enables one to evaluate the performance of a prediction algorithm. Our main aim in this paper is to propose a framework for model-based prediction methods that can be combined with sophisticated machine learning algorithms realizing the benefit of both model-based and model-free approaches. We demonstrated the performance of these methods using SARIMA models and neural networks, and identified possible directions for further improvement of these methods.   

\section*{Acknowledgments} We would like to thank  Pilot Travel Centers LLC and Sensys Networks, Inc. for sharing the data used in this study. We are grateful to Rahul Jain for his valuable comments. This research was supported by National Science Foundation EAGER award 1839843.

\bibliographystyle{unsrt}
\bibliography{mybibfile}

\end{document}